\documentclass{article}

\usepackage[final]{corl_2026} 
\usepackage{graphicx}
\usepackage{amsmath}
\usepackage{algorithm}
\usepackage[noend]{algpseudocode}
\usepackage{amssymb}
\usepackage{caption}
\usepackage{colortbl}
\usepackage{booktabs}
\usepackage{tabularx}
\usepackage{enumitem}
\usepackage{titletoc}

\usepackage{wrapfig}

\definecolor{mygray}{gray}{.85}
\definecolor{verylightgray}{RGB}{240,240,240} 

\title{F2F-AP: Flow-to-Future Asynchronous Policy for Real-time Dynamic Manipulation}

\author{
  Haoyu Wei$^{123}$, \space Xiuwei Xu$^{123}$\thanks{Project lead.$^\ddagger$ Corresponding author}\space,  \space Ziyang Cheng$^{123}$,  \space Hang Yin$^{123}$,  \space Angyuan Ma$^{123}$\\[3pt]
  \textbf{Jie Zhou}$^{123}$, \space \textbf{Jiwen Lu}$^{123}$, \space \textbf{Bingyao Yu}$^{123\ddagger}$ \\[3pt]
  $^1$Department of Automation, Tsinghua University \\[3pt]
  $^2$Beijing Key Laboratory of Embodied Intelligence Systems \\[3pt]
  $^3$Institute for Embodied Intelligence and Robotics, Tsinghua University
}

\begin{document}

\maketitle

\begin{abstract}
Asynchronous inference has emerged as a prevalent paradigm in robotic manipulation, achieving progress in ensuring trajectory smoothness and efficiency. However, a systemic challenge remains unresolved, as inherent latency causes generated actions to inevitably lag behind the real-time environment. This issue is particularly exacerbated in dynamic scenarios, where such temporal misalignment severely compromises the policy's ability to interpret and react to rapidly evolving surroundings. In this paper, we propose a novel framework that leverages predicted object flow to synthesize future observations, incorporating a flow-based contrastive learning objective to align the visual feature representations of predicted observations with ground-truth future states. Empowered by this anticipated visual context, our asynchronous policy gains the capacity for proactive planning and motion, enabling it to explicitly compensate for latency and robustly execute manipulation tasks involving actively moving objects. Experimental results demonstrate that our approach significantly enhances responsiveness and success rates in complex dynamic manipulation tasks. \href{https://why-peace.github.io/F2F-AP/}{Website}.
\end{abstract}
\vspace{-1.5mm}
\vspace{-1.5mm}

\section{Introduction}

Despite recent impressive advancements in robotic manipulation~\citep{intelligence2025,wu2025,xie2026}, the inherent computational latency of policy inference often hinders real-world deployment. Under straightforward synchronous execution strategies, robots frequently exhibit sluggish and discontinuous motion. Consequently, there is a growing trend towards adopting asynchronous inference frameworks, which parallelize model inference with action execution to significantly improve response speeds.

Action chunking~\citep{zhao2023,chi2024a,liao2025} serves as the foundation for asynchronous inference. Under this paradigm, the model predicts a sequence of actions covering multiple future timesteps, allowing the robot to infer the next action chunk during the execution of the current chunk. However, this introduces two critical issues. First, since chunked actions align with the start of inference, inference latency renders the initial actions outdated upon execution, forcing them to be discarded. Second, compounding latencies cause the policy to persistently lag behind the evolving environment and robot state. This problem is particularly acute in dynamic tasks. While static tasks might rely on a single optimal trajectory, dynamic scenarios involve highly complex and continuously shifting action spaces.

\begin{figure}
    \includegraphics[width=\textwidth]{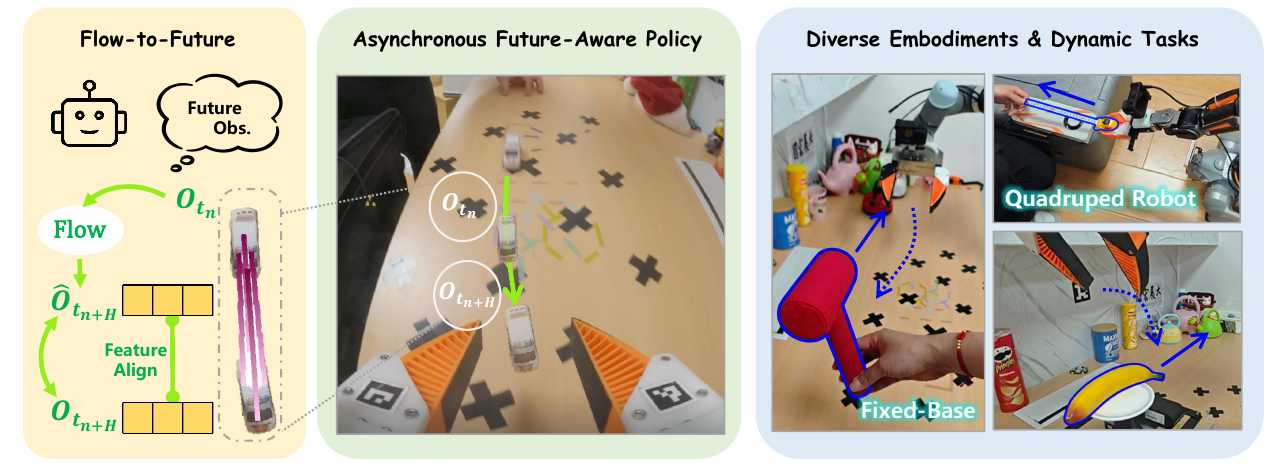}
    \captionof{figure}{\textbf{Flow-to-Future Asynchronous Policy (F2F-AP).} F2F-AP is an asynchronous robot manipulation policy incorporating future observations. By predicting object flow, it enhances the model's understanding of the motion trends of interacting objects. F2F-AP is transferable to diverse embodiments and yields significant performance improvements in dynamic tasks.}
    \vspace{-5mm}
    \label{fig:teaser}
\end{figure}

A representative recent work, VLASH~\citep{tang2025}, mitigates this by conditioning policies on anticipated future proprioceptive states. By predicting actions for the exact moment of execution, it eliminates initial chunk discards and smooths trajectory transitions. However, a critical flaw still remains: the temporal misalignment between stale visual observations and anticipated states. Without access to future visual cues, the policy is forced to ``guess" future actions based on insufficient, outdated visual information rather than accurately ``inferring" them from a synchronized real-time context. Consequently, achieving robust and accurate real-time dynamic manipulation with asynchronous policies remains an open challenge.

To address these challenges, we propose \textbf{Flow-to-Future Asynchronous Policy (F2F-AP)}, a future-aware framework tailored for real-time dynamic robotic manipulation, as shown in Figure~\ref{fig:teaser}. F2F-AP utilizes the predicted object flow of interacting objects as a dynamic scene representation to synthesize future visual observations. Furthermore, we propose a \textit{flow contrastive learning} objective to bridge the gap between predicted and ground-truth future observations, minimizing feature discrepancy to guarantee the fidelity of the anticipated context. By explicitly modeling systemic latency within an asynchronous pipeline, F2F-AP successfully generates action chunks that perfectly align with real-world timing, ensuring naturally connected trajectories. Simultaneously, the synthesized future observations effectively compensate for the information deficit caused by latency in dynamic tasks. Extensive experiments demonstrate that F2F-AP achieves significantly superior performance over prior methods across multiple dynamic tasks on two distinct hardware platforms, fully validating the necessity and effectiveness of our proposed design.

\section{Related Work}
\subsection{Affordance-based Manipulation}
Bridging the gap between raw sensory inputs and action outputs is a core challenge in robotic manipulation. To help policies implicitly learn intermediate knowledge like object localization and functionality, prior work heavily integrates affordances. These include object-centric instances~\citep{do2018a,mo2021,zhang2024,qian2024,bahl2023}, keypoints~\citep{gao2021,pan2023,fang2024,singh2025,girgin2025}, foundation models~\citep{huang,nasiriany2024,ju2024}, and implicit neural fields~\citep{simeonov2021,weng2023,chun2023}.

Most relevant to our approach are flow-based affordances. Flow, or point-trajectory, provides an embodiment-agnostic representation~\citep{yuan2024,xu2024,tang2025a,dharmarajan2025} that facilitates cross-domain data transfer, such as from human demonstrations to robots. Flow also serves as a high-level action interface~\citep{eisner2024,wen2024,bharadhwaj2025,zhi2025,zhang2025,huang2026}, guiding policy generation or enabling direct motion prediction~\citep{su2025}. However, these studies primarily focus on moving interactive objects within static environments, largely neglecting the spontaneous motion of the objects themselves. In contrast, F2F-AP extends flow utility to real-time dynamic scenarios by characterizing motion trends via object flow prediction and encoding future observational features.

\subsection{Asynchronous Inference}
Action chunking~\citep{zhao2023,chi2024a,liao2025} allows the robots to predict multiple future steps simultaneously, while asynchronous inference~\citep{shukor2025,sendai2025} parallelizes this prediction with execution to achieve higher frequencies and smoother deployment on physical systems. However, the asynchrony method frequently causes trajectory discontinuities. While existing methods~\citep{black,tang2025,wang2026,black2025} attempt
\begin{wrapfigure}{r}{0.6\linewidth}
  \centering
    \includegraphics[width=\linewidth]{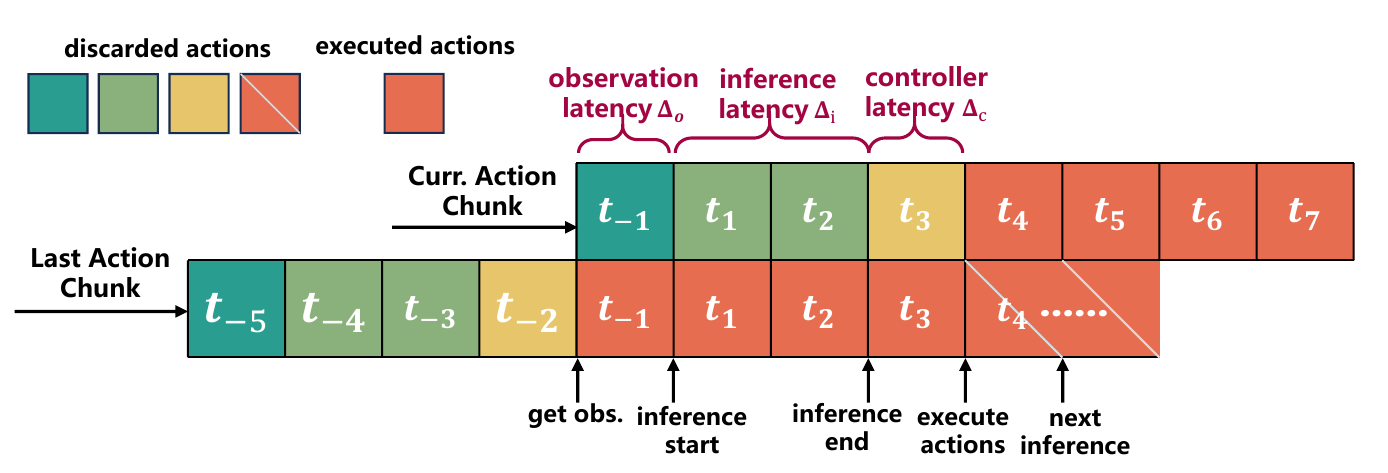}
  \caption{
  \textbf{Decomposition of systemic latency.} The total latency ($\Delta_o + \Delta_i + \Delta_c$) results in a misalignment where the initial frames of the predicted trajectory lag behind the real-world state, invalidating them for execution.
  }
  \vspace{-3mm}
  \label{fig:latency}
\end{wrapfigure}
to smooth transitions using trajectory inpainting, state-conditional inputs, or future states, they fail to address the intrinsic latency that causes actions to lag behind an evolving environment. F2F-AP resolves this via a future-observation synthesis module, empowering the policy to plan ahead and compensate for latency. This minimizes real-world trajectory redundancy and further boosts the efficiency of asynchronous inference.

\section{Method}

This section begins by clarifying the obstacles facing asynchronous policies. Next, we elaborate on our core contributions: the Flow2Future Module, Flow Contrastive Learning and the F2F policy.

\subsection{Problem Statement.}\label{sec:problem statement}
Asynchronous inference for dynamic tasks is challenging due to the systemic latency in the real world. As shown in Figure~\ref{fig:latency}, the total latency primarily stems from three components: \textit{Observation Latency~($\Delta_o$):} The time delay inherent in sensor acquisition and transmission, meaning the observation used for inference already lags behind the current state. \textit{Inference Latency~($\Delta_i$):} The computational time consumed by the model to generate output. \textit{Controller Latency~($\Delta_c$):} The delay introduced by the low-level controller during execution.

In summary, during the total systemic latency $\Delta_{\text{total}} = \Delta_o + \Delta_i + \Delta_c$, a robot operating under asynchronous inference continues to execute actions from the current chunk. We define the parameter $H$ to denote the number of action steps executed during this latency within a single cycle.

\begin{figure*}[t]
  \centering
    \includegraphics[width=\linewidth]{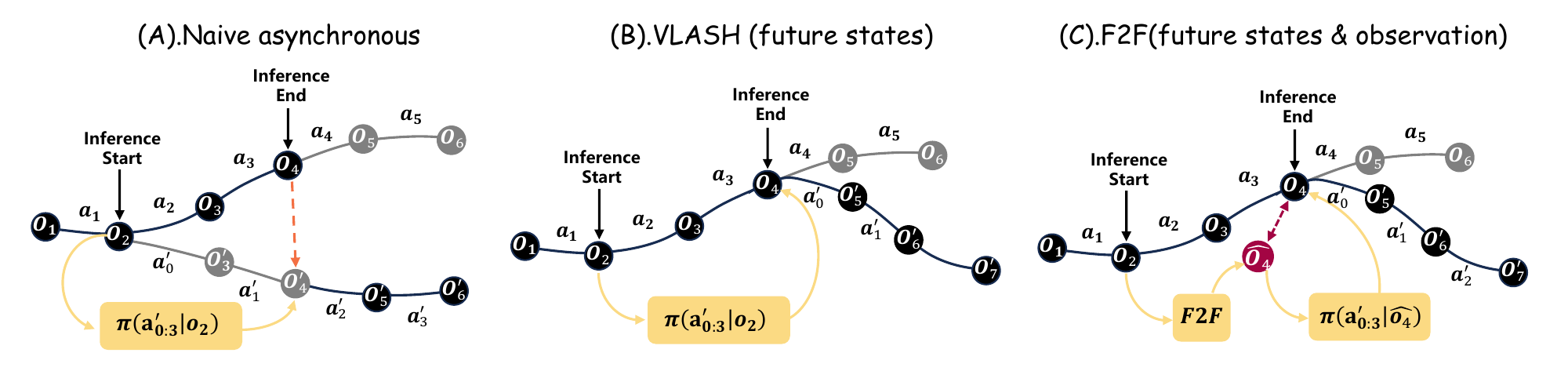}
  \caption{
      \textbf{Analysis of Asynchronous Inference.} Gray and black denote unexecuted and executed actions, respectively. \textbf{Left}: The orange line shows the trajectory gap. \textbf{Middle}: Policy input $o_2$ is misaligned with $o_4$ at execution time. \textbf{Right} (Ours): Uses predicted, aligned observation $\hat{\mathbf{o}_4}$.
  }
  \vspace{-3mm}
  \label{fig:asynchronous}
\end{figure*}

\textbf{Preliminary.}
A naive asynchronous policy can be formulated as~$\pi(\mathbf{a}_{t:t+n} | \mathbf{s}_t, \mathbf{o}_t)$, as illustrated in Figure~\ref{fig:asynchronous}. The policy utilizes the observation $\mathbf{o}_t$ and proprioceptive state $\mathbf{s}_t$ to predict actions starting from timestep $t$. However, due to systemic latency, the first $H$ predicted actions become outdated by the time inference completes, forcing them to be discarded. Furthermore, action mismatches within the overlapping temporal windows of consecutive chunks yield non-smooth trajectories, typically requiring action fusion~\citep{black}. This limitation is particularly detrimental in dynamic tasks.

Alternatively, VLASH~\citep{tang2025} proposes a new method from a training perspective, formulating it as~$\pi(\mathbf{a}_{t+H:t+H+n} | \mathbf{s}_{t+H}, \mathbf{o}_t)$. By utilizing the anticipated future state $\mathbf{s}_{t+H}$ as an anchor, VLASH directly outputs actions corresponding to the exact moment of execution. However, this paradigm introduces temporal misalignment between the lagging visual context $\mathbf{o}_t$ and the advanced state $\mathbf{s}_{t+H}$. This input imbalance makes it highly difficult for the policy to robustly respond to the rapidly evolving environments in dynamic interaction tasks.

\textbf{Overview.} 
F2F-AP aims to resolve the inherent latency problem in asynchronous policies by temporally aligning future proprioceptive states and visual observations, which can be formulated as~$\pi(\mathbf{a}_{t+H:t+H+n} | \mathbf{s}_{t+H}, \hat{\mathbf{o}}_{t+H})$. F2F-AP employs predicted object flow as a bridge to synthesize future observations. Furthermore, the action chunks generated by F2F-AP are strictly aligned with the exact moment of execution, effectively eliminating the need for discarding initial $H$ action steps or employing post-hoc action fusion algorithms. Through this design, F2F-AP substantially improves the model's performance on dynamic tasks.

\subsection{Flow2Future Module}

In order to predict the future observation, a naive approach would be to fine-tune off-the-shelf video generation models~\citep{kong2025,yang2025,wan2025} on the target tasks. However, current video generation models typically demand seconds of inference time per generation, rendering them entirely incapable of handling dynamic tasks that require rapid responses. 

In contrast, we propose the Flow2Future (F2F) module, which leverages predictive object flow as a bridge between current and future observations. The flow predictor employed in the F2F module provides an effective and significantly lightweight alternative, making it better suited for real-time deployment on physical robotic systems. The model architecture can be found in Appendix~\ref{sec:model_archtecture}.

\begin{figure*}[t]
  \centering
    \includegraphics[width=\linewidth]{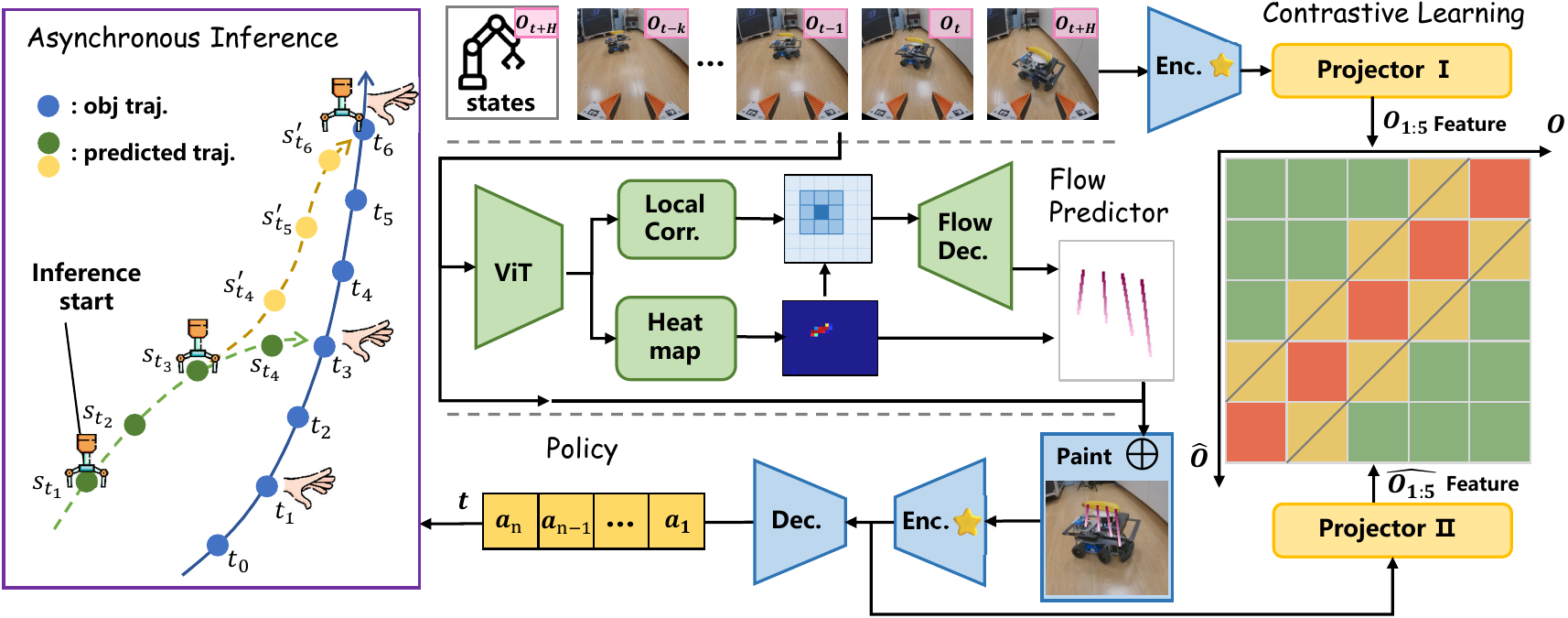}
  \caption{
      \textbf{(Left)} The model plans from a future state $s_{t_3}$ toward the interacting object's anticipated position $t_6$ at timestamp $t_1$, enabling advance motion despite systemic latency. \textbf{(Middle)} Given robot states and multi-frame RGB images, a \textit{Flow Predictor} extracts object flow to synthesize future observations. The \textit{Policy} then processes these to generate action chunks. \textbf{(Right)} Contrastive learning minimizes the feature distance between predicted and real future observations. The \textcolor[RGB]{241,191,51}{$\bigstar$} indicates features sharing the same encoder.
    }
  \vspace{-3mm}
  \label{fig:pipeline}
\end{figure*}

\textbf{Heatmap-Based Flow Predictor.}
F2F aims to construct a spatio-temporal visual model that predicts the motion trends of target objects through feature evolution, as shown in Figure~\ref{fig:pipeline}. However, the arbitrary indexing of ground truth points causes standard matching losses to disrupt feature mapping, forcing the model to converge to a trivial averaged flow, as detailed in Section~\ref{sec:model_archtecture}.

To address this, F2F adopts \textit{a multi-branch decoder centered on heatmap prediction} instead of naive regression. Formally, the object flow prediction $\mathbf{f}$ is defined as a composition of a refined starting position and a motion displacement vector:
\begin{equation} 
    \mathbf{f}_{t \to t+H} = \left[ \mathcal{S}(H(Z_v))_{\text{coarse}} + D_{\text{off}}(Z_v, Z_h)_{\text{fine}}, \quad D_{\mathbf{d}}(Z_v, Z_h)_{\text{motion}} \right]
    \label{eq:flow}
\end{equation}
Here, $\mathbf{f}_{t \to t+H}$ denotes the predicted object flow over a temporal horizon $H$. $Z_v$ represents image feature maps extracted from historical frames $o_{t-n \to t}$, and $H(\cdot)$ is the heatmap computation module. The soft-sampling operation $\mathcal{S}(\cdot)$ yields the \textit{coarse flow starting point} $\mathbf{f}_s$ by aggregating spatial coordinates weighted by the heatmap's response values. Using this, $Z_h$ is derived by computing local feature correlations across temporal frames. Finally, the flow decoder outputs $D_{\text{off}}$ and $D_{\mathbf{d}}$, which represent the fine-grained starting point refinement and the flow displacement vector, respectively.

The predicted flow is then utilized to synthesize the future observation:
\begin{equation}
    \hat{\mathbf{o}}_{t+H} = \mathcal{F}(\mathbf{o}_{t-n \to t}) = \mathrm{Paint}(\mathbf{f}_{t \to t+H}, \mathbf{o}_t)
    \label{eq:o_hat}
\end{equation}
where $\mathcal{F}(\cdot)$ denotes the F2F module. The operation $\mathrm{Paint}(\cdot)$ represents the process of rendering the flow vectors $\mathbf{f}_{t \to t+H}$ onto the current observation, where the vector is assigned a specific color to represent its magnitude and direction of displacement from its origin, shown in Section~\ref{sec:visualization_flow_prediction}.

\textbf{Data Collection.}
We built a data collection pipeline using the UMI~\citep{chi2024} device, which is capable of capturing trajectories, fisheye RGB images, and ground-truth object flow, even when the interactive object is occluded. Details are provided in Section~\ref{sec:data details}.

\subsection{Flow Contrastive Learning}

Although the predicted future observation $\hat{\mathbf{o}}_{t+H}$ derived from Equation~\ref{eq:o_hat} preserves comprehensive flow information, a naive concatenation of this prediction with the current observation may fail to yield a future-aware representation comparable to that of the real future observation $\mathbf{o}_{t+H}$. This representational discrepancy would confuse the policy and undermine our primary objective of synthesizing future observations for asynchronous inference.

To mitigate this, F2F incorporates contrastive learning~\citep{radford2021,zhang2026}, which aims to minimize the representation gap between flow-painted $\hat{\mathbf{o}}_{t+H}$ and real $\mathbf{o}_{t+H}$: $\mathop{\min}_{\theta} \mathrm{Diff}(\mathcal{E}_{\theta}(\hat{\mathbf{o}}_{t+H}), \mathcal{E}_{\theta}(\mathbf{o}_{t+H}))$ where $\mathcal{E}$ is the policy encoder and $\mathrm{Diff}$ indicates the difference in feature space.

In the contrastive learning module as illustrated in Figure~\ref{fig:pipeline}, predicted and ground-truth observations are mapped to a latent space using a shared encoder and distinct projectors. By computing their normalized cosine similarity, we maximize the representation alignment for synchronized pairs and minimize it for asynchronous ones.

However, adjacent frames share strong visual correlations, meaning that indiscriminately treating them as negative samples introduces false negatives that disrupt training. To mitigate this, we design a temporal mask $\mathbf{M}$ to ignore temporally proximal frames. With this mask applied, the contrastive loss $\ell(u, v)$ only optimizes synchronized positive pairs against sufficiently distant negative pairs:
\begin{equation}
\ell(u, v) = - \frac{1}{N} \sum_{i=1}^{N} \log \frac{e^{\langle u_i, v_i \rangle / \tau}}{\sum_{k=1}^{N} \mathbf{M}_{i,k} \cdot e^{\langle u_i, v_k \rangle / \tau}},
\text{ where }\mathbf{M}_{i,k} = \mathbb{I}(d(i, k) = 0 \lor d(i, k) \geq \tau_d)
\label{eq:directional_loss}
\end{equation}
where $\langle \cdot, \cdot \rangle$ denotes the cosine similarity, $d(i, k)$ is the temporal distance between frames and $\tau_d$ is a distance threshold hyperparameter. Letting $z^g = \mathcal{E}_{\theta}(o)$ and $z^p = \mathcal{E}_{\theta}(\hat{o})$ be the latent features of the ground-truth and predicted observations extracted by the shared encoder $\mathcal{E}_{\theta}$, our final symmetric objective is computed as $\mathcal{L} = \frac{1}{2} \left[ \ell(z^p, z^g) + \ell(z^g, z^p) \right]$. The visualization is in Section~\ref{sec:visualization_contrastive}.

\subsection{F2F Asynchronous Policy}

\textbf{Policy Architecture.}
The F2F policy is built upon the Diffusion Policy framework as detailed in Section~\ref{sec:policy architecture}. The overall policy can be expressed by the following formulation:
\begin{equation}
    \mathcal{P}(\hat{\mathbf{o}}_{t+H},\mathbf{s}_{t+H})=\mathcal{D} \circ \mathcal{E}(\mathcal{F}(\mathbf{o}_{t-n \to t}),\mathbf{s}_{t+H})
\end{equation}
where $\mathcal{D}$ denotes the decoder within the policy. As depicted in Figure~\ref{fig:pipeline}, F2F-AP synthesizes future observations via the flow predictor $\mathcal{F}$, followed by a Diffusion Policy to generate action trajectories.

\textbf{Explicit Latency Modeling.}
Given that the total latency of a specific system is bounded, setting the prediction horizon above this threshold ensures that all action timestamps fall beyond the latency period. F2F-AP calculates this required prediction horizon $H$ based on the total latency $\Delta_{\text{total}}$ and the control time step $\delta_t$: $H = \lceil \Delta_{\text{total}} / \delta_t \rceil$. In practice, individual latency components can be reliably estimated~\citep{chi2024}, enabling us to explicitly model the overall systemic latency.

\textbf{Action Schedule.} Since the predicted actions $a_{t+H \to t+H+n}$ bypass the latency and align explicitly with the real-world timeline, they can be executed directly without discarding initial steps or employing fusion algorithms. We simply maintain a queue for sequential execution, employing a low-level trajectory planner to ensure the actions are completed at their precise corresponding timestamps.

\section{Experiment}
\begin{figure*}[t]
  \centering
    \includegraphics[width=\linewidth]{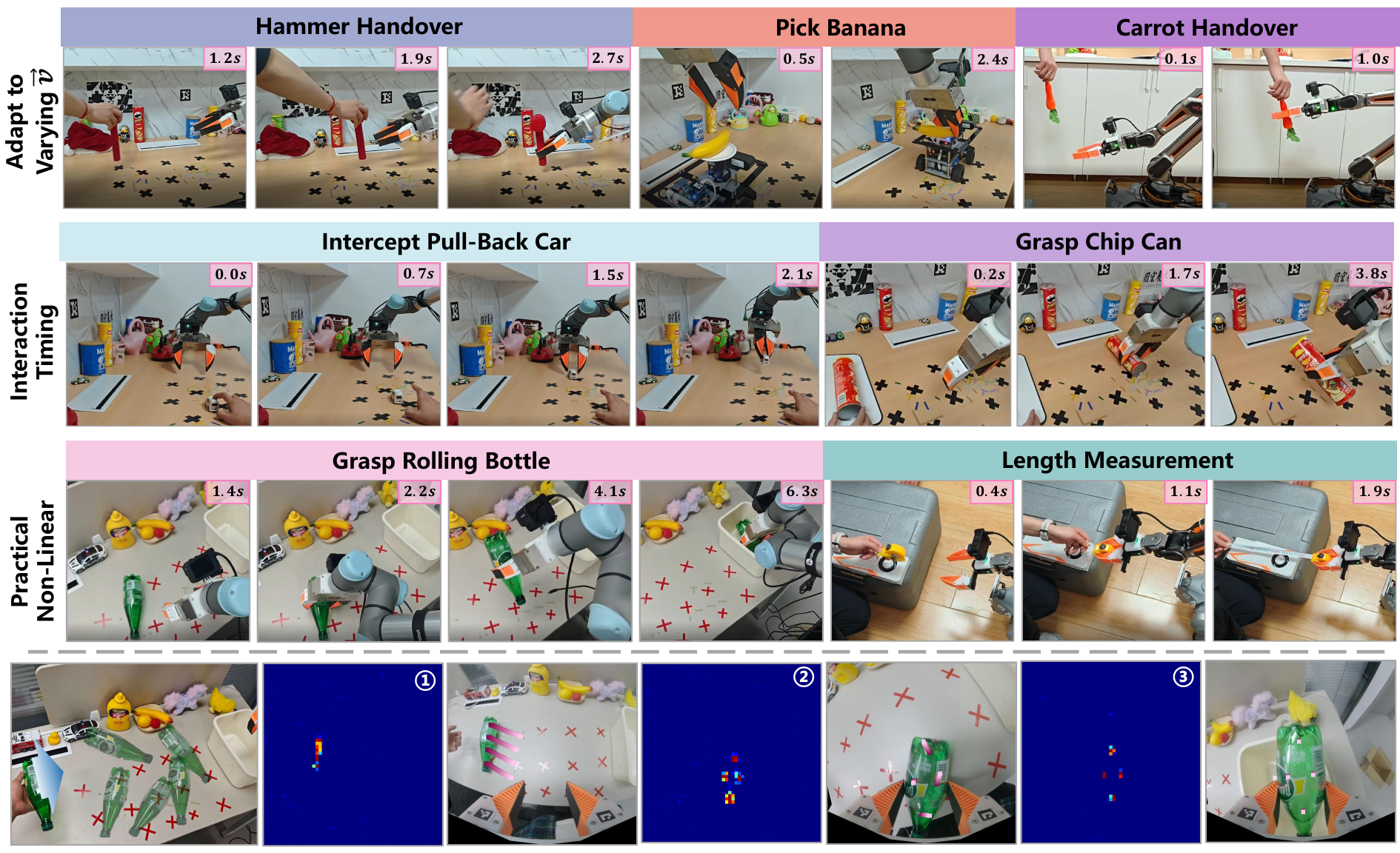}
  \caption{
  \textbf{Experiment Visualization.} \textbf{(Top)} Visualization of the real-world dynamic experiments. \textbf{(Bottom)} The non-linear stochastic trajectories, along with the heatmaps and predicted flow from the robot's first-person perspective.
  }
  \vspace{-3mm}
  \label{fig:vis_results}
\end{figure*}

To assess the capability of F2F-AP in dynamic scenarios, we designed seven tasks as illustrated in Figure~\ref{fig:vis_results}. Through comparative experiments for imitation learning policies, we demonstrate that F2F-AP significantly enhances policy performance in real-time dynamic tasks.

\subsection{Experiment Settings}
In this section, we detail the platform setup, the experimental baselines and evaluation metrics.

\textbf{Platforms.}
To demonstrate the versatility of our approach, we deploy it on two distinct robotic platforms, as detailed in Section~\ref{sec:platforms}, encompassing a UR5e fixed-base single-arm system and a quadruped mobile manipulator with a Whole-Body Controller (WBC). The quadruped experiments evaluate F2F-AP on an additional embodiment with the extra latency introduced by WBC and network communication.

In practice, the system latencies are measured to be $\Delta_o \approx 125\,\mathrm{ms}$, $\Delta_i \approx 200\,\mathrm{ms}$, and $\Delta_c < 50\,\mathrm{ms}$. Based on these measured delays and the action sampling interval $\delta_t = 0.1\,\mathrm{s}$, we determine the future horizon parameter $H = 4$ to ensure temporal alignment.

\textbf{Baseline methods.}
We select UMI~\citep{chi2024} for the fixed-base robotic arm and UMI-on-Legs~\citep{ha2024} for the quadruped mobile manipulator as our baselines and evaluate them under both synchronous and asynchronous inference settings. Additionally, we evaluate RTC~\citep{black} and VLASH~\citep{tang2025} as two advanced methods integrated into the UMI policy.

\textbf{Metrics.}
We employ \textit{Task Success Rate (SR)} and \textit{Average Execution Time (ET)} as two primary metrics to evaluate model performance, specifically focusing on reliability and efficiency. SR is defined as the ratio of successful trials to the total number, while ET denotes the mean duration of successful episodes, measured from the initiation of the first action to task termination.

\subsection{Capability Assessment and Tasks}
To characterize the dynamic response of F2F-AP, we focus on two dimensions: \textit{adaptability to varying velocities and directions of interacting objects}, and \textit{the temporal precision of the interaction timing}. Furthermore, we evaluate the model's performance in practical applications characterized by \textit{non-linear trajectories with high randomness}.

\textbf{Adaptability to Varying Velocities and Directions.}
A key feature of dynamic tasks is that interacting objects move at varying velocities and directions, requiring the model to possess the capacity to adapt to such uncertainties. We design three tasks for testing this capability.

\begin{itemize}[leftmargin=20pt,itemsep=3pt, parsep=0pt, topsep=0pt]
    \item \textbf{Human-to-Robot Hammer Handover.} This task requires the robot to swiftly receive a hammer from a human participant, who intentionally averts their gaze to ensure task stochasticity.
    \item \textbf{Pick Up a Moving Banana.} The robot is required to grasp a banana from a platform moving continuously forward at randomized speeds ranging from 0.1 m/s to 0.4 m/s.
    \item \textbf{Human-to-Robot Carrot Handover.} Similar to the hammer handover but using a quadruped manipulator instead of a single arm, this task requires the robot to intercept a carrot handed over by a human participant rapidly.
\end{itemize}

\textbf{Temporal Precision of Interaction Timing.}
The time window for completing dynamic tasks is typically narrow. Therefore, the robot must accurately determine the interaction timing and precisely execute the predicted actions in real-time. We evaluate this capability across two sub-second tasks.

\begin{itemize}[leftmargin=20pt,itemsep=3pt, parsep=0pt, topsep=0pt]
    \item \textbf{Intercept Pull-Back Car.} The robot is required to intercept a rapidly moving pull-back car, which features a restricted effective grasping region of 5 cm and an average velocity of 25 cm/s, resulting in a narrow timing window of under 0.4 s.
    \item \textbf{Grasp Rolling Chip-Can.} The objective is to grasp a chip can rolling down a ramp and subsequently return to its starting pose.
\end{itemize}

\textbf{Practical Non-Linear Tasks.}\label{part:rolling bottle}
Real-world dynamic tasks often feature non-linear trajectories and multiple motion stages, requiring models to adapt to continuously changing dynamic demands. We designed two practical tasks to validate the effectiveness of our approach.

\begin{itemize}[leftmargin=20pt,itemsep=3pt, parsep=0pt, topsep=0pt]
    \item \textbf{Grasp Rolling Bottle.} The robot needs to grasp a rolling, half-filled water bottle, which exhibits a highly stochastic and non-linear trajectory due to the randomness of human throws, its inherent asymmetry, and complex internal fluid dynamics.
    \item \textbf{Collaborative Length Measurement.} The task requires the quadruped robot to sequentially execute three actions: grasping the tape measure immediately upon the handover, pressing it down firmly, and waiting for the human measurement without jitter.
\end{itemize}

\begin{table}[t]
    \centering
    \caption{
        \textbf{Quantitative Results.} Comparison against baselines in real-world experiments. Length Meas. denotes Collaborative Length Measurement.
    }
    \label{tab:quantitative results}
    \small
    \setlength{\tabcolsep}{3pt}
    \resizebox{\linewidth}{!}{
        \begin{tabularx}{\textwidth}{@{}l *{14}{>{\centering\arraybackslash}X}@{}} 
            \toprule
            \textbf{Method} 
            & \multicolumn{2}{c}{\textbf{Hammer}}
            & \multicolumn{2}{c}{\textbf{Banana}} 
            & \multicolumn{2}{c}{\textbf{Carrot}} 
            & \multicolumn{2}{c}{\makebox[0pt][c]{\textbf{Pull-Back Car}}}
            & \multicolumn{2}{c}{\textbf{Chip Can}} 
            & \multicolumn{2}{c}{\textbf{Bottle}} 
            & \multicolumn{2}{c}{\makebox[0pt][c]{\textbf{Length Meas.}}} \\
            
            \cmidrule(lr){2-3} \cmidrule(lr){4-5} \cmidrule(lr){6-7} \cmidrule(lr){8-9} \cmidrule(lr){10-11} \cmidrule(lr){12-13} \cmidrule(lr){14-15}
            
            & \textbf{SR} & \textbf{ET} & \textbf{SR} & \textbf{ET} & \textbf{SR} & \textbf{ET} & \textbf{SR} & \textbf{ET} & \textbf{SR} & \textbf{ET} & \textbf{SR} & \textbf{ET} & \textbf{SR} & \textbf{ET} \\
            \midrule
            
            Syn.  UMI & 5/20 & 5.7s & 5/20 & 3.0s & 3/10 & 2.8s & 0/20 & - & 3/20 & 4.1s & 2/20 & 9.2s & 4/10 & 2.6s\\
            Asyn. UMI & 7/20 & 6.2s & 10/20 & 2.9s & 5/10 & 2.4s & 1/20 & 1.4s & 2/20 & 4.2s & 2/20 & 9.1s & 5/10 & 2.4s\\
            RTC & 12/20 & 4.1s  & 12/20 & 2.4s & 6/20 & 2.0s & 0/20 & - & 3/20 & 3.9s & 3/20 & 8.2s & 5/20 & 2.2s\\
            VLASH & 6/20 & 3.3s & 13/20 & 2.5s & 3/10 & 1.4s & 3/20 & 1.9s & 1/20 & 3.0s & 2/20 & 7.6s & 3/10 & 1.4s\\
            \rowcolor{mygray} \textbf{F2F-AP} & 15/20 & 3.4s & 13/20 & 2.5s & 6/10 & 1.6s & 8/20 & 2.5s & 9/20 & 3.5s & 11/20 & 7.9s & 7/10 & 1.9s\\
            \bottomrule
        \end{tabularx}
    }
    \vspace{-5mm}
\end{table}

\textbf{Results Analysis.}
The quantitative results of the seven experiments are presented in Table~\ref{tab:quantitative results}. The quantitative results indicate that F2F-AP leads in overall success rate. In highly dynamic tasks such as the \textit{hammer handover}, \textit{intercepting the pull-back car}, and \textit{grasping the rolling bottle}, F2F-AP achieves more than a twofold increase in success rate. Regarding execution time, F2F-AP significantly outperforms UMI, achieving efficiency comparable to VLASH and RTC.

Based on experimental observations, synchronous UMI shows slow responsiveness, whereas asynchronous UMI suffers from delayed, jittery trajectories with repetitive attempts. Although RTC improves trajectory smoothness, it cannot cope with highly stochastic tasks. Despite generating smoother and faster movements than UMI, VLASH fails to meet real-world temporal requirements, frequently overshooting or entirely missing the critical interaction timing.

The inherent latency within asynchronous architectures is the main bottleneck for dynamic tasks. As shown in Section~\ref{sec:problem statement}, the lagging context impairs the model's capability to respond to dynamic changes in an evolving environment. Additionally, data collection pipelines implicitly ignore most latency, introducing a training-inference gap. F2F-AP effectively overcomes these limitations by predicting future observations and temporally aligning them with proprioceptive states, ensuring that inferred actions are strictly executed at the precise real-world moments.

\subsection{Ablation Study}\label{sec:ablation}
To further validate the efficacy of our policy design, we conducted ablation studies targeting key aspects, with results summarized in Table~\ref{tab:ablation}.

\begin{itemize}[leftmargin=20pt,itemsep=3pt, parsep=0pt, topsep=0pt]
    \item \textbf{Hyperparameters.} We modified the flow color from pink to green and flow points from four to two. The negligible performance impact demonstrates the robustness of F2F-AP.
    \item \textbf{Key Components.} We conduct ablation studies on flow contrastive learning, the use of future proprioceptive states, and the heatmap in flow prediction. The removal of any of these components degrades the model's performance, underscoring the necessity of our design.
    \item \textbf{History Information.} We adjusted the number of RGB input frames for UMI to match that of F2F-AP's flow predictor, and conducted experiments both with and without RTC. The near-complete failure of Multi-Frames UMI proves that F2F-AP's superiority arises not from additional historical context, but from its robust adaptability to dynamics and latency.
    \item \textbf{Diff. Latency.} We observe that a lower inference frequency leads to degraded performance. Furthermore, we test F2F-AP with varying latency, demonstrating robustness across different latency levels, experiencing a significant performance drop only under excessively high latency.
\end{itemize}

\begin{table}[t]
    \centering
    \small
    \caption{
        \textbf{Ablation Studies.} The results reported correspond to the \textit{Grasp Rolling Bottle} task (Section~\ref{part:rolling bottle}).
    }
    \label{tab:ablation}
    \setlength{\tabcolsep}{6pt} 
    \begin{tabularx}{0.9\linewidth}{>{\raggedright\arraybackslash}X c c | >{\raggedright\arraybackslash}X c c}
        \toprule
        \textbf{Method} & \textbf{SR} & \textbf{ET} (s) & \textbf{Method} & \textbf{SR} & \textbf{ET} (s)\\
        \midrule
        Fewer Flow Points    & 10/20 & 7.8  & MF-UMI              & 1/20  & 8.7  \\
        Diff. Line Color     & 11/20 & 7.8  & RTC-MF-UMI          & 0/20  & -    \\
        Low Contrastive Wt.  & 7/20  & 8.5  & Low Freq. Inference & 3/20  & 10.2 \\
        w/o Contrastive Loss & 8/20  & 8.5  & 300ms Latency       & 12/20 & 7.5  \\
        w/o Future State     & 3/20  & 8.3  & 500ms Latency       & 10/20 & 8.5  \\
        w/o Heatmap          & 5/20  & 9.1  & 600ms Latency       & 4/20  & 11.2 \\
        \rowcolor{mygray} \textbf{Full F2F-AP} & \textbf{11/20} & \textbf{7.9} & \textbf{Full F2F-AP} & \textbf{11/20} & \textbf{7.9}  \\
        \bottomrule
    \end{tabularx}
    \vspace{-5mm}
\end{table}

\section{Conclusion and Limitation}\label{sec:conclusion}
In this work, we introduce the Flow-to-Future Asynchronous Policy (F2F-AP) to address the critical challenge of delayed decision-making in dynamic environments. Within an asynchronous inference framework, F2F-AP incorporates the Flow2Future module to estimate object flow and directly synthesize future observations, enabling the generation of future-aware action chunks. We developed a heatmap-based model to stabilize flow prediction and employed flow contrastive learning to minimize the feature discrepancy between predicted and ground-truth observations. Extensive experiments across two distinct robotic platforms demonstrate that F2F-AP significantly enhances success rates in dynamic tasks, reduces overall execution time, and improves system responsiveness.

\textbf{Limitation and Future Works.} While F2F-AP empowers the policy with future-aware capabilities, several avenues remain for future exploration. First, excessive latency uncertainty caused by factors such as network fluctuations would make the current fixed-horizon approach difficult to apply. To resolve this, future work will focus on developing a model capable of forecasting flow across varying time horizons. Second, to overcome the current limitation of single-task scenarios, we plan to design a universal, task-conditioned flow predictor that enables the scaling toward generalist models. Finally, we will extend the F2F-AP framework from 2D pixel space to support 3D inputs, lifting representations into real-world coordinates to endow the model with superior spatial understanding.

\acknowledgments{This work was supported in part by the National Key Research and Development Program of China under Grant 2022ZD0114903,  in part by the National Natural Science Foundation of China under Grant 62336004, in part by Beijing Natural Science Foundation under Grant L247009, and in part by the Beijing Major Science and Technology Project under Contract  no. Z251100008125019 and no. Z251100008125064.}


\bibliography{main}  

@inproceedings{bahl2023,
  booktitle = {2023 IEEE/CVF Conference on Computer Vision and Pattern Recognition (CVPR)},
  author = {Bahl, Shikhar and Mendonca, Russell and Chen, Lili and Jain, Unnat and Pathak, Deepak},
  year = 2023,
  month = jun,
  pages = {01--13},
  publisher = {IEEE},
  address = {Vancouver, BC, Canada},
  urldate = {2026-01-27},
  copyright = {https://doi.org/10.15223/policy-029},
  isbn = {979-8-3503-0129-8},
  langid = {english},
  title = {\href{https://ieeexplore.ieee.org/document/10204422/}{Affordances from Human Videos as a Versatile Representation for Robotics}}
}

@inproceedings{bharadhwaj2025,
  title = {\href{https://arxiv.org/abs/2405.01527}{Track2Act: Predicting Point Tracks from~Internet Videos Enables Generalizable Robot Manipulation}},
  shorttitle = {Track2Act},
  booktitle = {Computer Vision -- ECCV 2024},
  author = {Bharadhwaj, Homanga and Mottaghi, Roozbeh and Gupta, Abhinav and Tulsiani, Shubham},
  editor = {Leonardis, Ale{\v s} and Ricci, Elisa and Roth, Stefan and Russakovsky, Olga and Sattler, Torsten and Varol, G{\"u}l},
  year = 2025,
  pages = {306--324},
  publisher = {Springer Nature Switzerland},
  address = {Cham},
  isbn = {978-3-031-73116-7},
  langid = {english}
}

@article{black,
  title = {\href{https://arxiv.org/abs/2506.07339}{Real-Time Execution of Action Chunking Flow Policies}},
  author = {Black, Kevin and Galliker, Manuel Y and Levine, Sergey},
  langid = {english},
  journal={arXiv preprint arXiv:2506.07339},
}

@article{black2025,
  author = {Black, Kevin and Ren, Allen Z. and Equi, Michael and Levine, Sergey},
  year = 2025,
  month = dec,
  journal = {arXiv:2512.05964},
  eprint = {2512.05964},
  primaryclass = {cs},
  publisher = {arXiv},
  urldate = {2025-12-12},
  archiveprefix = {arXiv},
  title = {\href{http://arxiv.org/abs/2512.05964}{Training-Time Action Conditioning for Efficient Real-Time Chunking}}
}

@article{chi2024,
  shorttitle = {Universal Manipulation Interface},
  author = {Chi, Cheng and Xu, Zhenjia and Pan, Chuer and Cousineau, Eric and Burchfiel, Benjamin and Feng, Siyuan and Tedrake, Russ and Song, Shuran},
  year = 2024,
  month = jan,
  journal = {CoRR},
  urldate = {2025-05-11},
  langid = {english},
  title = {\href{https://openreview.net/forum?id=Kcak3D6ZX6}{Universal Manipulation Interface: In-The-Wild Robot Teaching Without In-The-Wild Robots}}
}

@article{chi2024a,
  shorttitle = {Diffusion Policy},
  author = {Chi, Cheng and Xu, Zhenjia and Feng, Siyuan and Cousineau, Eric and Du, Yilun and Burchfiel, Benjamin and Tedrake, Russ and Song, Shuran},
  year = 2024,
  month = mar,
  journal = {arXiv:2303.04137},
  eprint = {2303.04137},
  primaryclass = {cs},
  publisher = {arXiv},
  urldate = {2025-05-16},
  archiveprefix = {arXiv},
  title = {\href{http://arxiv.org/abs/2303.04137}{Diffusion Policy: Visuomotor Policy Learning via Action Diffusion}}
}

@inproceedings{chun2023,
  shorttitle = {Local Neural Descriptor Fields},
  booktitle = {2023 IEEE International Conference on Robotics and Automation (ICRA)},
  author = {Chun, Ethan and Du, Yilun and Simeonov, Anthony and {Lozano-Perez}, Tomas and Kaelbling, Leslie},
  year = 2023,
  month = may,
  pages = {1830--1836},
  urldate = {2026-01-27},
  title = {\href{https://ieeexplore.ieee.org/document/10160423/}{Local Neural Descriptor Fields: Locally Conditioned Object Representations for Manipulation}}
}

@article{dharmarajan2025,
  shorttitle = {Dream2Flow},
  author = {Dharmarajan, Karthik and Huang, Wenlong and Wu, Jiajun and {Fei-Fei}, Li and Zhang, Ruohan},
  year = 2025,
  month = dec,
  journal = {arXiv:2512.24766},
  eprint = {2512.24766},
  primaryclass = {cs},
  publisher = {arXiv},
  urldate = {2026-01-20},
  archiveprefix = {arXiv},
  langid = {american},
  title = {\href{http://arxiv.org/abs/2512.24766}{Dream2Flow: Bridging Video Generation and Open-World Manipulation with 3D Object Flow}}
}

@inproceedings{do2018a,
  shorttitle = {AffordanceNet},
  booktitle = {2018 IEEE International Conference on Robotics and Automation (ICRA)},
  author = {Do, Thanh-Toan and Nguyen, Anh and Reid, Ian},
  year = 2018,
  month = may,
  pages = {5882--5889},
  issn = {2577-087X},
  urldate = {2026-01-27},
  title = {\href{https://ieeexplore.ieee.org/document/8460902/}{AffordanceNet: An End-to-End Deep Learning Approach for Object Affordance Detection}}
}

@article{dong2024,
  shorttitle = {MemFlow},
  author = {Dong, Qiaole and Fu, Yanwei},
  year = 2024,
  month = apr,
  journal = {arXiv:2404.04808},
  eprint = {2404.04808},
  primaryclass = {cs},
  publisher = {arXiv},
  urldate = {2025-08-12},
  archiveprefix = {arXiv},
  langid = {american},
  title = {\href{http://arxiv.org/abs/2404.04808}{MemFlow: Optical Flow Estimation and Prediction with Memory}},
  journal={arXiv preprint arXiv:2404.04808},
}

@article{eisner2024,
  shorttitle = {FlowBot3D},
  author = {Eisner, Ben and Zhang, Harry and Held, David},
  year = 2024,
  month = may,
  journal = {arXiv:2205.04382},
  eprint = {2205.04382},
  primaryclass = {cs},
  publisher = {arXiv},
  urldate = {2026-01-29},
  archiveprefix = {arXiv},
  langid = {english},
  title = {\href{http://arxiv.org/abs/2205.04382}{FlowBot3D: Learning 3D Articulation Flow to Manipulate Articulated Objects}}
}

@inproceedings{fang2024,
  shorttitle = {MOKA},
  booktitle = {Robotics: Science and Systems XX},
  author = {Fang, Kuan and Liu, Fangchen and Abbeel, Pieter and Levine, Sergey},
  year = 2024,
  month = jul,
  publisher = {{Robotics: Science and Systems Foundation}},
  urldate = {2026-01-27},
  isbn = {979-8-9902848-0-7},
  langid = {english},
  title = {\href{http://www.roboticsproceedings.org/rss20/p062.pdf}{MOKA: Open-World Robotic Manipulation through Mark-Based Visual Prompting}}
}

@article{gao2021,
  shorttitle = {kPAM 2.0},
  author = {Gao, Wei and Tedrake, Russ},
  year = 2021,
  month = apr,
  journal = {IEEE Robotics and Automation Letters},
  volume = {6},
  number = {2},
  pages = {2962--2969},
  issn = {2377-3766},
  urldate = {2026-01-27},
  title = {\href{https://ieeexplore.ieee.org/document/9363532/}{kPAM 2.0: Feedback Control for Category-Level Robotic Manipulation}}
}

@article{girgin2025,
  shorttitle = {Multiobject Graph Affordance Network},
  author = {Girgin, Tuba and U{\u g}ur, Emre},
  year = 2025,
  month = aug,
  journal = {IEEE Transactions on Cognitive and Developmental Systems},
  volume = {17},
  number = {4},
  pages = {847--858},
  issn = {2379-8939},
  urldate = {2026-01-27},
  title = {\href{https://ieeexplore.ieee.org/document/10807281/}{Multiobject Graph Affordance Network: Goal-Oriented Planning Through Learned Compound Object Affordances}}
}

@inproceedings{ha2024,
  shorttitle = {UMI on Legs},
  booktitle = {CoRL 2024 Workshop on Whole-Body Control and Bimanual Manipulation: Applications in Humanoids and Beyond},
  author = {Ha, Huy and Gao, Yihuai and Fu, Zipeng and Tan, Jie and Song, Shuran},
  year = 2024,
  month = nov,
  urldate = {2025-05-11},
  langid = {english},
  title = {\href{https://openreview.net/forum?id=VtJMXwHj5p}{UMI on Legs: Making Manipulation Policies Mobile with Manipulation-Centric Whole-body Controllers}}
}

@article{huang,
  title = {\href{https://arxiv.org/abs/2307.05973}{VoxPoser: Composable 3D Value Maps for Robotic Manipulation with Language Models}},
  author = {Huang, Wenlong and Wang, Chen and Zhang, Ruohan and Li, Yunzhu and Wu, Jiajun and {Fei-Fei}, Li},
  langid = {english},
  journal={arXiv preprint arXiv:2307.05973},
  year={2023}
}

@article{huang2026,
  shorttitle = {PointWorld},
  author = {Huang, Wenlong and Chao, Yu-Wei and Mousavian, Arsalan and Liu, Ming-Yu and Fox, Dieter and Mo, Kaichun and {Fei-Fei}, Li},
  year = 2026,
  month = jan,
  journal = {arXiv:2601.03782},
  eprint = {2601.03782},
  primaryclass = {cs},
  publisher = {arXiv},
  urldate = {2026-01-18},
  archiveprefix = {arXiv},
  langid = {american},
  title = {\href{http://arxiv.org/abs/2601.03782}{PointWorld: Scaling 3D World Models for In-The-Wild Robotic Manipulation}}
}

@article{intelligence2025,
  shorttitle = {\${$\pi\_$}\textbraceleft 0.5\textbraceright\$},
  author = {Intelligence, Physical and Black, Kevin and Brown, Noah and Darpinian, James and Dhabalia, Karan and Driess, Danny and Esmail, Adnan and Equi, Michael and Finn, Chelsea and Fusai, Niccolo and Galliker, Manuel Y. and Ghosh, Dibya and Groom, Lachy and Hausman, Karol and Ichter, Brian and Jakubczak, Szymon and Jones, Tim and Ke, Liyiming and LeBlanc, Devin and Levine, Sergey and {Li-Bell}, Adrian and Mothukuri, Mohith and Nair, Suraj and Pertsch, Karl and Ren, Allen Z. and Shi, Lucy Xiaoyang and Smith, Laura and Springenberg, Jost Tobias and Stachowicz, Kyle and Tanner, James and Vuong, Quan and Walke, Homer and Walling, Anna and Wang, Haohuan and Yu, Lili and Zhilinsky, Ury},
  year = 2025,
  month = apr,
  journal = {arXiv:2504.16054},
  eprint = {2504.16054},
  primaryclass = {cs},
  publisher = {arXiv},
  urldate = {2025-05-15},
  archiveprefix = {arXiv},
  title = {\href{http://arxiv.org/abs/2504.16054}{$\pi_{0.5}$: a Vision-Language-Action Model with Open-World Generalization}}
}

@article{karaev2024cotracker3simplerbetter,
  shorttitle = {CoTracker3},
  author = {Karaev, Nikita and Makarov, Iurii and Wang, Jianyuan and Neverova, Natalia and Vedaldi, Andrea and Rupprecht, Christian},
  year = 2024,
  month = oct,
  journal = {arXiv:2410.11831},
  eprint = {2410.11831},
  primaryclass = {cs},
  publisher = {arXiv},
  urldate = {2026-01-31},
  archiveprefix = {arXiv},
  title = {\href{http://arxiv.org/abs/2410.11831}{CoTracker3: Simpler and Better Point Tracking by Pseudo-Labelling Real Videos}}
}

@article{ju2024,
  shorttitle = {Robo-ABC},
  author = {Ju, Yuanchen and Hu, Kaizhe and Zhang, Guowei and Zhang, Gu and Jiang, Mingrun and Xu, Huazhe},
  year = 2024,
  month = jan,
  journal = {arXiv:2401.07487},
  eprint = {2401.07487},
  primaryclass = {cs},
  publisher = {arXiv},
  urldate = {2026-01-27},
  archiveprefix = {arXiv},
  title = {\href{http://arxiv.org/abs/2401.07487}{Robo-ABC: Affordance Generalization Beyond Categories via Semantic Correspondence for Robot Manipulation}}
}

@article{kong2025,
  shorttitle = {HunyuanVideo},
  author = {Kong, Weijie and Tian, Qi and Zhang, Zijian and Min, Rox and Dai, Zuozhuo and Zhou, Jin and Xiong, Jiangfeng and Li, Xin and Wu, Bo and Zhang, Jianwei and Wu, Kathrina and Lin, Qin and Yuan, Junkun and Long, Yanxin and Wang, Aladdin and Wang, Andong and Li, Changlin and Huang, Duojun and Yang, Fang and Tan, Hao and Wang, Hongmei and Song, Jacob and Bai, Jiawang and Wu, Jianbing and Xue, Jinbao and Wang, Joey and Wang, Kai and Liu, Mengyang and Li, Pengyu and Li, Shuai and Wang, Weiyan and Yu, Wenqing and Deng, Xinchi and Li, Yang and Chen, Yi and Cui, Yutao and Peng, Yuanbo and Yu, Zhentao and He, Zhiyu and Xu, Zhiyong and Zhou, Zixiang and Xu, Zunnan and Tao, Yangyu and Lu, Qinglin and Liu, Songtao and Zhou, Dax and Wang, Hongfa and Yang, Yong and Wang, Di and Liu, Yuhong and Jiang, Jie and Zhong, Caesar},
  year = 2025,
  month = mar,
  journal = {arXiv:2412.03603},
  eprint = {2412.03603},
  primaryclass = {cs},
  publisher = {arXiv},
  urldate = {2026-01-31},
  archiveprefix = {arXiv},
  title = {\href{http://arxiv.org/abs/2412.03603}{HunyuanVideo: A Systematic Framework For Large Video Generative Models}}
}

@article{liao2025,
  shorttitle = {Delay-Aware Diffusion Policy},
  author = {Liao, Aileen and Kim, Dong-Ki and Smith, Max Olan and {Agha-mohammadi}, Ali-akbar and Omidshafiei, Shayegan},
  year = 2025,
  month = dec,
  journal = {arXiv:2512.07697},
  eprint = {2512.07697},
  primaryclass = {cs},
  publisher = {arXiv},
  urldate = {2025-12-19},
  archiveprefix = {arXiv},
  langid = {american},
  title = {\href{http://arxiv.org/abs/2512.07697}{Delay-Aware Diffusion Policy: Bridging the Observation-Execution Gap in Dynamic Tasks}}
}

@inproceedings{mo2021,
  shorttitle = {Where2Act},
  booktitle = {Proceedings of the IEEE/CVF International Conference on Computer Vision},
  author = {Mo, Kaichun and Guibas, Leonidas J. and Mukadam, Mustafa and Gupta, Abhinav and Tulsiani, Shubham},
  year = 2021,
  pages = {6813--6823},
  urldate = {2026-01-27},
  langid = {english},
  title = {\href{https://openaccess.thecvf.com/content/ICCV2021/html/Mo_Where2Act_From_Pixels_to_Actions_for_Articulated_3D_Objects_ICCV_2021_paper.html}{Where2Act: From Pixels to Actions for Articulated 3D Objects}}
}

@article{nasiriany2024,
  shorttitle = {RT-Affordance},
  author = {Nasiriany, Soroush and Kirmani, Sean and Ding, Tianli and Smith, Laura and Zhu, Yuke and Driess, Danny and Sadigh, Dorsa and Xiao, Ted},
  year = 2024,
  month = nov,
  journal = {arXiv:2411.02704},
  eprint = {2411.02704},
  primaryclass = {cs},
  publisher = {arXiv},
  urldate = {2026-01-27},
  archiveprefix = {arXiv},
  langid = {english},
  title = {\href{http://arxiv.org/abs/2411.02704}{RT-Affordance: Affordances are Versatile Intermediate Representations for Robot Manipulation}}
}

@inproceedings{pan2023,
  shorttitle = {TAX-Pose},
  booktitle = {Proceedings of The 6th Conference on Robot Learning},
  author = {Pan, Chuer and Okorn, Brian and Zhang, Harry and Eisner, Ben and Held, David},
  year = 2023,
  month = mar,
  pages = {1783--1792},
  publisher = {PMLR},
  issn = {2640-3498},
  urldate = {2026-01-27},
  langid = {english},
  title = {\href{https://proceedings.mlr.press/v205/pan23a.html}{TAX-Pose: Task-Specific Cross-Pose Estimation for Robot Manipulation}}
}

@inproceedings{qian2024,
  shorttitle = {AffordanceLLM},
  booktitle = {2024 IEEE/CVF Conference on Computer Vision and Pattern Recognition Workshops (CVPRW)},
  author = {Qian, Shengyi and Chen, Weifeng and Bai, Min and Zhou, Xiong and Tu, Zhuowen and Li, Li Erran},
  year = 2024,
  month = jun,
  pages = {7587--7597},
  publisher = {IEEE},
  address = {Seattle, WA, USA},
  urldate = {2026-01-27},
  copyright = {https://doi.org/10.15223/policy-029},
  isbn = {979-8-3503-6547-4},
  langid = {english},
  title = {\href{https://ieeexplore.ieee.org/document/10678001/}{AffordanceLLM: Grounding Affordance from Vision Language Models}}
}

@inproceedings{radford2021,
  booktitle = {Proceedings of the 38th International Conference on Machine Learning},
  author = {Radford, Alec and Kim, Jong Wook and Hallacy, Chris and Ramesh, Aditya and Goh, Gabriel and Agarwal, Sandhini and Sastry, Girish and Askell, Amanda and Mishkin, Pamela and Clark, Jack and Krueger, Gretchen and Sutskever, Ilya},
  year = 2021,
  month = jul,
  pages = {8748--8763},
  publisher = {PMLR},
  issn = {2640-3498},
  urldate = {2026-01-31},
  langid = {english},
  title = {\href{https://proceedings.mlr.press/v139/radford21a.html}{Learning Transferable Visual Models From Natural Language Supervision}}
}

@article{ravi2024a,
  shorttitle = {SAM 2},
  author = {Ravi, Nikhila and Gabeur, Valentin and Hu, Yuan-Ting and Hu, Ronghang and Ryali, Chaitanya and Ma, Tengyu and Khedr, Haitham and R{\"a}dle, Roman and Rolland, Chloe and Gustafson, Laura and Mintun, Eric and Pan, Junting and Alwala, Kalyan Vasudev and Carion, Nicolas and Wu, Chao-Yuan and Girshick, Ross and Doll{\'a}r, Piotr and Feichtenhofer, Christoph},
  year = 2024,
  month = oct,
  journal = {arXiv:2408.00714},
  eprint = {2408.00714},
  primaryclass = {cs},
  publisher = {arXiv},
  urldate = {2026-01-27},
  archiveprefix = {arXiv},
  langid = {american},
  title = {\href{http://arxiv.org/abs/2408.00714}{SAM 2: Segment Anything in Images and Videos}}
}

@article{sendai2025,
  shorttitle = {Leave No Observation Behind},
  author = {Sendai, Kohei and Alvarez, Maxime and Matsushima, Tatsuya and Matsuo, Yutaka and Iwasawa, Yusuke},
  year = 2025,
  month = sep,
  journal = {arXiv:2509.23224},
  eprint = {2509.23224},
  primaryclass = {cs},
  publisher = {arXiv},
  urldate = {2026-01-29},
  archiveprefix = {arXiv},
  langid = {american},
  title = {\href{http://arxiv.org/abs/2509.23224}{Leave No Observation Behind: Real-time Correction for VLA Action Chunks}}
}

@article{shukor2025,
  shorttitle = {SmolVLA},
  author = {Shukor, Mustafa and Aubakirova, Dana and Capuano, Francesco and Kooijmans, Pepijn and Palma, Steven and Zouitine, Adil and Aractingi, Michel and Pascal, Caroline and Russi, Martino and Marafioti, Andres and Alibert, Simon and Cord, Matthieu and Wolf, Thomas and Cadene, Remi},
  year = 2025,
  month = jun,
  journal = {arXiv:2506.01844},
  eprint = {2506.01844},
  primaryclass = {cs},
  publisher = {arXiv},
  urldate = {2026-01-29},
  archiveprefix = {arXiv},
  langid = {american},
  title = {\href{http://arxiv.org/abs/2506.01844}{SmolVLA: A Vision-Language-Action Model for Affordable and Efficient Robotics}}
}

@article{simeonov2021,
  shorttitle = {Neural Descriptor Fields},
  author = {Simeonov, Anthony and Du, Yilun and Tagliasacchi, Andrea and Tenenbaum, Joshua B. and Rodriguez, Alberto and Agrawal, Pulkit and Sitzmann, Vincent},
  year = 2021,
  month = dec,
  journal = {arXiv:2112.05124},
  eprint = {2112.05124},
  primaryclass = {cs},
  publisher = {arXiv},
  urldate = {2026-01-27},
  archiveprefix = {arXiv},
  title = {\href{http://arxiv.org/abs/2112.05124}{Neural Descriptor Fields: SE(3)-Equivariant Object Representations for Manipulation}}
}

@article{singh2025,
  shorttitle = {AFFORD2ACT},
  author = {Singh, Anukriti and Torshizi, Kasra and Habib, Khuzema and Yu, Kelin and Gao, Ruohan and Tokekar, Pratap},
  year = 2025,
  month = oct,
  journal = {arXiv:2510.01433},
  eprint = {2510.01433},
  primaryclass = {cs},
  publisher = {arXiv},
  urldate = {2026-01-27},
  archiveprefix = {arXiv},
  title = {\href{http://arxiv.org/abs/2510.01433}{AFFORD2ACT: Affordance-Guided Automatic Keypoint Selection for Generalizable and Lightweight Robotic Manipulation}}
}

@article{su2025,
  shorttitle = {Motion Before Action},
  author = {Su, Yue and Zhan, Xinyu and Fang, Hongjie and Li, Yong-Lu and Lu, Cewu and Yang, Lixin},
  year = 2025,
  month = apr,
  journal = {arXiv:2411.09658},
  eprint = {2411.09658},
  primaryclass = {cs},
  publisher = {arXiv},
  urldate = {2025-09-18},
  archiveprefix = {arXiv},
  langid = {american},
  title = {\href{http://arxiv.org/abs/2411.09658}{Motion Before Action: Diffusing Object Motion as Manipulation Condition}}
}

@article{tai2025,
  author = {Tai, Wei-En and Shih, Yu-Lin and Sun, Cheng and Wang, Yu-Chiang Frank and Chen, Hwann-Tzong},
  year = 2025,
  month = mar,
  journal = {arXiv:2503.06261},
  eprint = {2503.06261},
  primaryclass = {cs},
  publisher = {arXiv},
  urldate = {2025-11-27},
  archiveprefix = {arXiv},
  langid = {american},
  title = {\href{http://arxiv.org/abs/2503.06261}{Segment Anything, Even Occluded}}
}

@article{tang2025,
  shorttitle = {VLASH},
  author = {Tang, Jiaming and Sun, Yufei and Zhao, Yilong and Yang, Shang and Lin, Yujun and Zhang, Zhuoyang and Hou, James and Lu, Yao and Liu, Zhijian and Han, Song},
  year = 2025,
  month = nov,
  journal = {arXiv:2512.01031},
  eprint = {2512.01031},
  primaryclass = {cs},
  publisher = {arXiv},
  urldate = {2025-12-12},
  archiveprefix = {arXiv},
  langid = {american},
  title = {\href{http://arxiv.org/abs/2512.01031}{VLASH: Real-Time VLAs via Future-State-Aware Asynchronous Inference}}
}

@inproceedings{tang2025a,
  booktitle = {2025 IEEE International Conference on Robotics and Automation (ICRA)},
  author = {Tang, Weiliang and Pan, Jia-Hui and Zhan, Wei and Zhou, Jianshu and Yao, Huaxiu and Liu, Yun-Hui and Tomizuka, Masayoshi and Ding, Mingyu and Fu, Chi-Wing},
  year = 2025,
  month = may,
  pages = {2086--2093},
  urldate = {2026-01-29},
  title = {\href{https://ieeexplore.ieee.org/document/11127873/}{Embodiment-agnostic Action Planning via Object-Part Scene Flow}}
}

@article{tran,
  title = {\href{https://arxiv.org/abs/2210.06323}{AISFormer: Amodal Instance Segmentation with Transformer}},
  author = {Tran, Minh},
  langid = {english},
  journal={arXiv preprint arXiv:2210.06323},
}

@article{wan2025,
  shorttitle = {Wan},
  author = {Wan, Team and Wang, Ang and Ai, Baole and Wen, Bin and Mao, Chaojie and Xie, Chen-Wei and Chen, Di and Yu, Feiwu and Zhao, Haiming and Yang, Jianxiao and Zeng, Jianyuan and Wang, Jiayu and Zhang, Jingfeng and Zhou, Jingren and Wang, Jinkai and Chen, Jixuan and Zhu, Kai and Zhao, Kang and Yan, Keyu and Huang, Lianghua and Feng, Mengyang and Zhang, Ningyi and Li, Pandeng and Wu, Pingyu and Chu, Ruihang and Feng, Ruili and Zhang, Shiwei and Sun, Siyang and Fang, Tao and Wang, Tianxing and Gui, Tianyi and Weng, Tingyu and Shen, Tong and Lin, Wei and Wang, Wei and Wang, Wei and Zhou, Wenmeng and Wang, Wente and Shen, Wenting and Yu, Wenyuan and Shi, Xianzhong and Huang, Xiaoming and Xu, Xin and Kou, Yan and Lv, Yangyu and Li, Yifei and Liu, Yijing and Wang, Yiming and Zhang, Yingya and Huang, Yitong and Li, Yong and Wu, You and Liu, Yu and Pan, Yulin and Zheng, Yun and Hong, Yuntao and Shi, Yupeng and Feng, Yutong and Jiang, Zeyinzi and Han, Zhen and Wu, Zhi-Fan and Liu, Ziyu},
  year = 2025,
  month = apr,
  journal = {arXiv:2503.20314},
  eprint = {2503.20314},
  primaryclass = {cs},
  publisher = {arXiv},
  urldate = {2026-01-31},
  archiveprefix = {arXiv},
  title = {\href{http://arxiv.org/abs/2503.20314}{Wan: Open and Advanced Large-Scale Video Generative Models}}
}

@article{wang2026,
  author = {Wang, Haoxuan and Zhang, Gengyu and Yan, Yan and Shang, Yuzhang and Kompella, Ramana Rao and Liu, Gaowen},
  year = 2026,
  month = jan,
  journal = {arXiv:2601.20130},
  eprint = {2601.20130},
  primaryclass = {cs},
  publisher = {arXiv},
  urldate = {2026-01-31},
  archiveprefix = {arXiv},
  title = {\href{http://arxiv.org/abs/2601.20130}{Real-Time Robot Execution with Masked Action Chunking}}
}

@article{wen2024,
  author = {Wen, Chuan and Lin, Xingyu and So, John and Chen, Kai and Dou, Qi and Gao, Yang and Abbeel, Pieter},
  year = 2024,
  month = jul,
  journal = {arXiv:2401.00025},
  eprint = {2401.00025},
  primaryclass = {cs},
  publisher = {arXiv},
  urldate = {2025-02-26},
  archiveprefix = {arXiv},
  langid = {american},
  title = {\href{http://arxiv.org/abs/2401.00025}{Any-point Trajectory Modeling for Policy Learning}}
}

@inproceedings{weng2023,
  booktitle = {2023 IEEE International Conference on Robotics and Automation (ICRA)},
  author = {Weng, Thomas and Held, David and Meier, Franziska and Mukadam, Mustafa},
  year = 2023,
  month = may,
  pages = {1814--1821},
  urldate = {2026-01-27},
  title = {\href{https://ieeexplore.ieee.org/document/10160217/}{Neural Grasp Distance Fields for Robot Manipulation}}
}

@inproceedings{wu2025,
  shorttitle = {MoManipVLA},
  booktitle = {Proceedings of the Computer Vision and Pattern Recognition Conference},
  author = {Wu, Zhenyu and Zhou, Yuheng and Xu, Xiuwei and Wang, Ziwei and Yan, Haibin},
  year = 2025,
  pages = {1714--1723},
  urldate = {2025-06-24},
  langid = {english},
  title = {\href{https://openaccess.thecvf.com/content/CVPR2025/html/Wu_MoManipVLA_Transferring_Vision-language-action_Models_for_General_Mobile_Manipulation_CVPR_2025_paper.html}{MoManipVLA: Transferring Vision-language-action Models for General Mobile Manipulation}}
}

@article{xie2026,
  shorttitle = {DynamicVLA},
  author = {Xie, Haozhe and Wen, Beichen and Zheng, Jiarui and Chen, Zhaoxi and Hong, Fangzhou and Diao, Haiwen and Liu, Ziwei},
  year = 2026,
  month = jan,
  journal = {arXiv:2601.22153},
  eprint = {2601.22153},
  primaryclass = {cs},
  publisher = {arXiv},
  urldate = {2026-01-31},
  archiveprefix = {arXiv},
  title = {\href{http://arxiv.org/abs/2601.22153}{DynamicVLA: A Vision-Language-Action Model for Dynamic Object Manipulation}}
}

@article{xu2024,
  author = {Xu, Mengda and Xu, Zhenjia and Xu, Yinghao and Chi, Cheng and Wetzstein, Gordon and Veloso, Manuela and Song, Shuran},
  year = 2024,
  month = oct,
  journal = {arXiv:2407.15208},
  eprint = {2407.15208},
  primaryclass = {cs},
  publisher = {arXiv},
  urldate = {2025-08-05},
  archiveprefix = {arXiv},
  langid = {american},
  title = {\href{http://arxiv.org/abs/2407.15208}{Flow as the Cross-Domain Manipulation Interface}}
}

@article{yang2025,
  shorttitle = {CogVideoX},
  author = {Yang, Zhuoyi and Teng, Jiayan and Zheng, Wendi and Ding, Ming and Huang, Shiyu and Xu, Jiazheng and Yang, Yuanming and Hong, Wenyi and Zhang, Xiaohan and Feng, Guanyu and Yin, Da and Zhang, Yuxuan and Wang, Weihan and Cheng, Yean and Xu, Bin and Gu, Xiaotao and Dong, Yuxiao and Tang, Jie},
  year = 2025,
  month = mar,
  journal = {arXiv:2408.06072},
  eprint = {2408.06072},
  primaryclass = {cs},
  publisher = {arXiv},
  urldate = {2026-01-31},
  archiveprefix = {arXiv},
  title = {\href{http://arxiv.org/abs/2408.06072}{CogVideoX: Text-to-Video Diffusion Models with An Expert Transformer}}
}

@inproceedings{yuan2024,
  booktitle = {8th Annual Conference on Robot Learning},
  author = {Yuan, Chengbo and Wen, Chuan and Zhang, Tong and Gao, Yang},
  year = 2024,
  month = sep,
  urldate = {2026-01-29},
  langid = {english},
  title = {\href{https://openreview.net/forum?id=nmEt0ci8hi}{General Flow as Foundation Affordance for Scalable Robot Learning}}
}

@inproceedings{zhang2024,
  shorttitle = {SAM-E},
  booktitle = {Proceedings of the 41st International Conference on Machine Learning},
  author = {Zhang, Junjie and Bai, Chenjia and He, Haoran and Wang, Zhigang and Zhao, Bin and Li, Xiu and Li, Xuelong},
  year = 2024,
  month = jul,
  pages = {58579--58598},
  publisher = {PMLR},
  issn = {2640-3498},
  urldate = {2026-01-27},
  langid = {english},
  title = {\href{https://proceedings.mlr.press/v235/zhang24c.html}{SAM-E: Leveraging Visual Foundation Model with Sequence Imitation for Embodied Manipulation}}
}

@article{zhang2025,
  author = {Zhang, Fan and Gienger, Michael},
  year = 2025,
  month = nov,
  journal = {arXiv:2409.01083},
  eprint = {2409.01083},
  primaryclass = {cs},
  publisher = {arXiv},
  urldate = {2026-01-18},
  archiveprefix = {arXiv},
  langid = {american},
  title = {\href{http://arxiv.org/abs/2409.01083}{Affordance-based Robot Manipulation with Flow Matching}}
}

@article{zhang2026,
  shorttitle = {CLAP},
  author = {Zhang, Chubin and Wang, Jianan and Gao, Zifeng and Su, Yue and Dai, Tianru and Zhou, Cai and Lu, Jiwen and Tang, Yansong},
  year = 2026,
  month = jan,
  journal = {arXiv:2601.04061},
  eprint = {2601.04061},
  primaryclass = {cs},
  publisher = {arXiv},
  urldate = {2026-01-20},
  archiveprefix = {arXiv},
  langid = {english},
  title = {\href{http://arxiv.org/abs/2601.04061}{CLAP: Contrastive Latent Action Pretraining for Learning Vision-Language-Action Models from Human Videos}}
}

@inproceedings{zhao2023,
  booktitle = {ICML Workshop on New Frontiers in Learning, Control, and Dynamical Systems},
  author = {Zhao, Tony Z. and Kumar, Vikash and Levine, Sergey and Finn, Chelsea},
  year = 2023,
  month = jul,
  urldate = {2026-01-29},
  langid = {english},
  title = {\href{https://openreview.net/forum?id=e8Eu1lqLaf}{Learning Fine-Grained Bimanual Manipulation with Low-Cost Hardware}}
}

@article{zhi2025,
  shorttitle = {3DFlowAction},
  author = {Zhi, Hongyan and Chen, Peihao and Zhou, Siyuan and Dong, Yubo and Wu, Quanxi and Han, Lei and Tan, Mingkui},
  year = 2025,
  month = jun,
  journal = {arXiv:2506.06199},
  eprint = {2506.06199},
  primaryclass = {cs},
  publisher = {arXiv},
  urldate = {2025-09-11},
  archiveprefix = {arXiv},
  langid = {english},
  title = {\href{http://arxiv.org/abs/2506.06199}{3DFlowAction: Learning Cross-Embodiment Manipulation from 3D Flow World Model}}
}

\clearpage
\section*{Appendix}

Please visit our project website to view videos: \href{https://f2f-ap.github.io/}{website}.

\startcontents[appendices]
\printcontents[appendices]{l}{0}{\setcounter{tocdepth}{2}}
\clearpage

\appendix
\clearpage

\section{Hardware Platforms}\label{sec:platforms}
\begin{wrapfigure}{l}{0.5\linewidth}
  \centering
    \includegraphics[width=\linewidth]{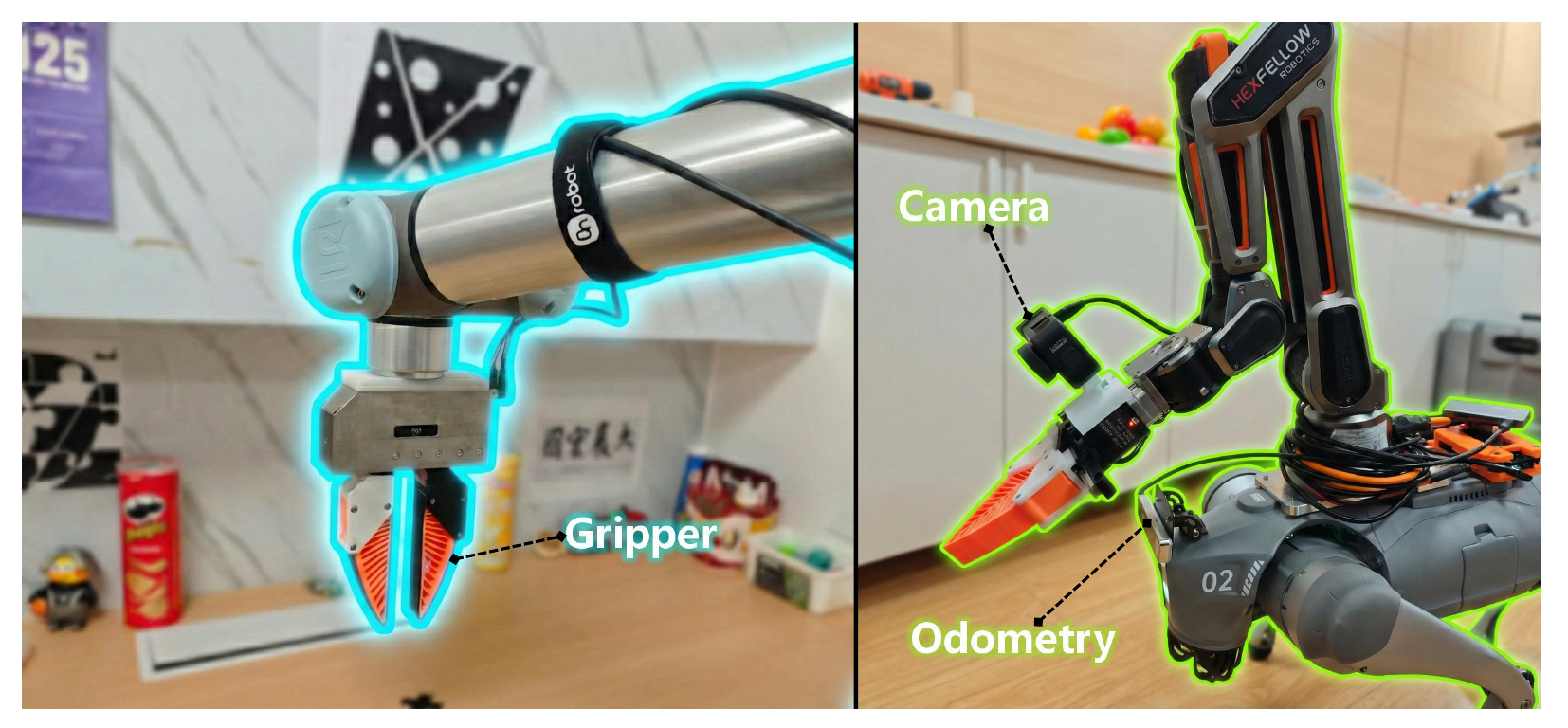}
  \caption{
  \textbf{Hardware Platform.} F2F-AP is evaluated on a fixed-base arm and a quadruped manipulator equipped with camera and odometry.
  }
  \label{fig:system}
\end{wrapfigure}

We evaluate the capability of F2F-AP on two different hardware platforms, as illustrated in Figure~\ref{fig:system}.

\textbf{Fixed-Base Robotic Arm.} Adhering to standard experimental configurations, we deploy the model on a static single-arm system. Specifically, we utilize the UR5e robotic arm for these experiments.

\textbf{Quadruped Mobile Manipulator.} Addressing the challenges of mobile manipulation, we employ a Unitree Go2 quadruped robot integrated with a Hexfellow Saber robotic arm. This system is controlled by a robust Whole-Body Controller (WBC), which provides unified control over the joints of both the quadruped base and the arm to precisely track end-effector pose trajectories.

\section{Data Details}\label{sec:data details}

\subsection{Data Collection Overview}
We utilize the Universal Manipulation Interface (UMI), equipped with a GoPro fisheye camera and a gripper, as the primary hardware for rapid data collection. For each single-task policy, we manually collect $100 \sim 200$ trajectories. Remarkably, only about 10 minutes of total video duration is required to train both the policy and the Flow-to-Future (F2F) module to achieve strong performance. This \textit{minimal data requirement allows for highly efficient data acquisition} and makes the system amenable to utilizing easily collected or internet-sourced videos. For fair comparison, all models are trained on the same source dataset collected in a real-world environment.

\subsection{Ground Truth Flow Generation.}
The fisheye images collected by UMI serve as the foundational training data for the flow predictor. To extract the ground truth object flow, we evaluated several methods—including point tracking~\citep{karaev2024cotracker3simplerbetter}, optical flow estimation~\citep{dong2024}, and filtering—and found the SAM~\citep{ravi2024a} image segmentation approach to be the most reliable for fisheye images. 

Consequently, we designed a segmentation-and-clustering pipeline, outlined in Algorithm~\ref{alg:flow_generation}. Following a human-annotated initial bounding box, SAM tracks the object mask across the video sequence. We then apply K-Means clustering ($K=4$) to extract representative keypoints on the interacting object. To ensure \textit{temporal consistency}, the clustering centroids for each frame are initialized using the keypoints from the previous frame. The ground truth flow displacement is subsequently computed as the pixel-wise distance of these keypoints across the prediction horizon $H$.

\begin{algorithm}[h]
\caption{Ground Truth Flow Generation Pipeline}
\label{alg:flow_generation}
\begin{algorithmic}[1]
    \Require RGB Video Sequence $\mathcal{V} = \{I_1, \dots, I_T\}$, Prediction Horizon $H$
    \Ensure Flow Ground Truth $\mathcal{G}_{flow}$
    
    \State \textbf{Initialization:} Human annotates bounding box $\mathbf{b}_1$ on $I_1$
    \State \Comment{Obtain object masks via SAM tracking}
    \State $\mathcal{M}_{1:T} \leftarrow \text{SAM-Track}(\mathcal{V}, \mathbf{b}_1)$ 
    
    \State Initialize keypoints set $\mathcal{P}_{0} \leftarrow \emptyset$
    \State \Comment{Extract temporal-consistent keypoints}
    \For{$t = 1$ to $T$}
        \If{$t = 1$}
            \State Init centroids $\mathbf{c}_{init}$ randomly within $\mathcal{M}_1$
        \Else
            \State Init centroids $\mathbf{c}_{init} \leftarrow \mathcal{P}_{t-1}$ \Comment{Use previous points as initialization}
        \EndIf
        \State $\mathcal{P}_t \leftarrow \text{K-Means}(\mathcal{M}_t, K=4, \text{init}=\mathbf{c}_{init})$
    \EndFor
    
    \State \Comment{Compute flow labels}
    \State $\mathcal{G}_{flow} \leftarrow \emptyset$
    \For{$t = 1$ to $T - H$}
        \State $\mathcal{P}_{future} \leftarrow \mathcal{P}_{t+H}$
        \State $\Delta \mathcal{P}_t \leftarrow \mathcal{P}_{future} - \mathcal{P}_t$ \Comment{Calculate displacement}
        \State $GT_t \leftarrow \{ (\mathbf{p}, \mathbf{d}) \mid \mathbf{p} \in \mathcal{P}_t, \mathbf{d} \in \Delta \mathcal{P}_t \}$
        \State $\mathcal{G}_{flow} \leftarrow \mathcal{G}_{flow} \cup GT_t$
    \EndFor
    
    \State \Return $\mathcal{G}_{flow}$
\end{algorithmic}
\end{algorithm}

\subsection{Handling Occlusions via Data Synthesis}
A significant challenge in pick-and-place tasks is severe object occlusion caused by the gripper (e.g., a teapot held by the gripper completely obscuring a cup during a pouring task). Existing segmentation and amodal~\citep{tran} models struggle to process such occluded objects in fisheye images. 

\begin{figure}
    \centering
    \includegraphics[width=\linewidth]{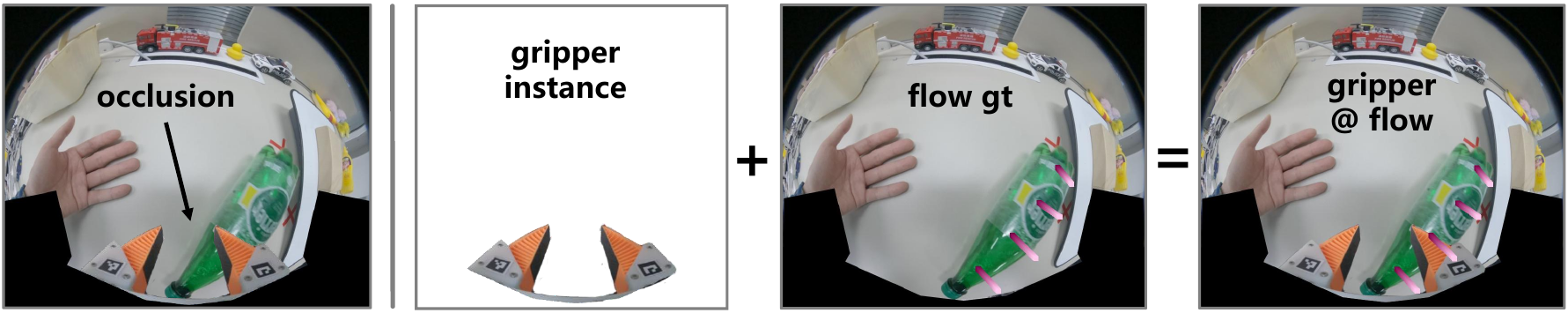}
    \caption{The left panel shows the occlusion scenarios, and the right panel illustrates how to build the ground-truth flow under occlusion. We select a pre-segmented gripper instance and overlay it onto the original fisheye videos to generate the RGB ground truth.}
    \label{fig:data_collection}
\end{figure}

To enable reliable flow prediction under severe occlusion, we propose a data synthesis approach inspired by SAMEO and following~\citep{tai2025}. Alongside the standard UMI dataset, we capture a separate, equivalently sized dataset of gripper-free video sequences using only the fisheye camera. These unobstructed videos allow us to obtain clean, accurate ground truth flow using the pipeline described above. We then synthesize the final training observations by randomly superimposing pre-segmented gripper masks onto these pure fisheye videos, effectively mimicking the original UMI captures. This synthesis process is illustrated in Figure~\ref{fig:data_collection}.

\section{Model Architecture}\label{sec:model_archtecture}

\subsection{Why Use a Heatmap?} 
During the training of the sparse object flow predictor, matching the predicted flows with the ground truth for loss computation presents a critical challenge. The ordering of the ground truth elements lacks physical significance, resulting in arbitrary indexing without fixed correspondence across different samples. For instance, a flow point tracking a bottle's neck might be indexed first in Episode A but last in Episode B. Consequently, applying standard one-to-one or minimum-distance losses—such as enforcing a strict index correspondence or utilizing the Hungarian algorithm—would disrupt the mapping between visual features and spatial coordinates, ultimately forcing the model to converge to an averaged flow as a trivial solution.

\subsection{Flow2Future Module}
The Flow2Future (F2F) module operates on an input video sequence $V \in \mathbb{R}^{6 \times 224 \times 224 \times 3}$. Initially, each frame is partitioned into $16 \times 16$ non-overlapping patches and linearly projected to an embedding dimension of $D=384$, augmented with learnable 2D spatial and 1D temporal sinusoidal positional embeddings. These tokenized representations are then processed by the backbone, which is an 8-layer BERT-style Transformer encoder. Specifically, the encoder is configured with 8 attention heads, a hidden dimension of 384, and an MLP intermediate dimension of 1536, utilizing a Pre-Norm structure with GELU activation. 

For flow estimation, the HeatmapFlowDecoder first computes a local correlation volume within a search radius of $r=2$ between the current and all of the past frame features. Subsequently, this volume is processed by a heatmap prediction branch—comprising a standard convolutional block followed by transposed convolution layers —to output a $56 \times 56$ probability heatmap. Upon extracting the keypoints via weighted K-Means, the corresponding features are sampled and fed into the flow regression MLP, which outputs the final sub-pixel offsets and flow vectors.

\subsection{Policy}\label{sec:policy architecture}
Our policy architecture adopts the UMI framework~\citep{chi2024}, utilizing a CLIP-pretrained ViT-Base encoder for observation encoding and a 1D Conditional U-Net for action generation. The policy operates on flow-augmented RGB images. Specifically, for each timestep in the observation horizon ($T_{obs}=2$), the frozen Flow2Future module predicts flow based on a history of $T_{pred}=6$ frames and overlays the resulting flow vectors onto the visual input. We employ a DDIM scheduler with 16 inference steps for the diffusion generation process.

\section{Implementation Details}
\subsection{Training}
Firstly, the Flow2Future Module is trained on eight 4090 GPUs using the Adam optimizer with a batch size of 32 and a base learning rate of $1 \times 10^{-4}$. We utilize a cosine annealing schedule with a 5-epoch warmup and enable mixed-precision training. The training objective combines an auxiliary image reconstruction loss, which encourages the model to capture global spatial context by restoring masked patches.

Then the policy is trained using the AdamW optimizer with a batch size of 32. We employ a cosine schedule with a 2000-step warmup, setting the base learning rate to $3 \times 10^{-4}$ while applying a reduced rate of $3 \times 10^{-5}$ to the vision backbone. The training objective incorporates a standard denoising loss and an auxiliary flow-contrastive loss—a key innovation of F2F-AP designed to align the embedding of the current observation with that of the future observation.

\subsection{Inference}
We execute our model on an NVIDIA RTX 3090 GPU with an inference frequency of 5 Hz, a standard rate for vision-based robotic manipulation. Meanwhile, the Whole-Body Controller of the quadruped mobile manipulator operates on an onboard NVIDIA Jetson Orin edge device.

\begin{figure}[t]
    \centering
    \includegraphics[width=\linewidth]{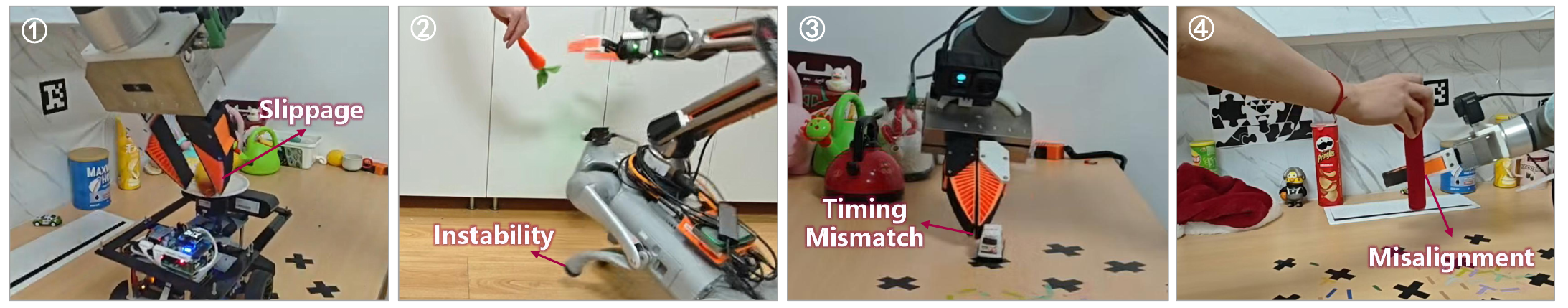}
    \caption{
    \textbf{Failure Case Visualization} illustrates four representative failure scenarios. 
    }
    \label{fig:failure_case}
    \vspace{-5mm}
\end{figure}

\begin{wrapfigure}{r}{0.5\linewidth}
    \centering
    \includegraphics[width=\linewidth]{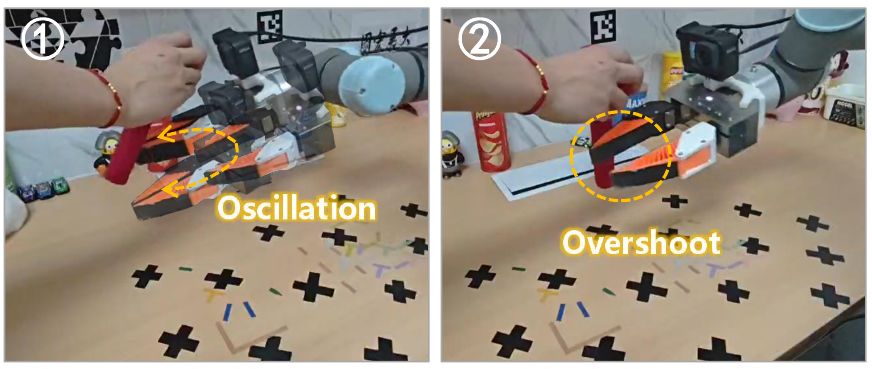}
    \caption{
    \textbf{Comparative Visualization} of baseline failure modes: (1) Naive Asynchronous Inference, and (2) VLASH.
    }
    \label{fig:compare}
\end{wrapfigure}

\section{Extended Evaluation}

\subsection{Comparative Analysis}
We documented the failure modes of Naive Asynchronous Inference and VLASH to provide a comparison with F2F-AP. Notably, all three methods were evaluated at the same inference frequency. As illustrated in Figure~\ref{fig:compare}, Naive Asynchronous Inference fails to generate temporally coherent actions due to systemic latency, manifesting as persistent grasping attempts that consistently miss the precise target location. In contrast, VLASH, lacking future observations, fails to anticipate the optimal stopping position in time, frequently resulting in overshooting.

\subsection{Failure Case Analysis}
We categorize the four typical failure modes encountered during the deployment of F2F-AP, as illustrated in Figure~\ref{fig:failure_case}:

\begin{itemize}[leftmargin=20pt,itemsep=3pt, parsep=0pt, topsep=0pt]
    \item \textbf{Object Slippage.} Suboptimal grasping poses often result in insecure contact, causing the object to be ejected or slip out of the gripper during the closing phase.
    \item \textbf{System Instability.} Interacting with dynamic objects introduces significant kinetic disturbances. When coupled with the inherent instability of the hardware, especially for the Whole-Body Controller of the quadruped manipulator, rapid posture adjustments can lead to the loss of balance or system tipping.
    \item \textbf{Timing Mismatch.} For dynamic tasks constrained by narrow interaction windows, errors in gripper actuation constitute the primary cause of task failure.
    \item \textbf{Spatial Misalignment.} The policy may occasionally fail to achieve precise spatial alignment with the target interaction point, resulting in the end-effector missing the object entirely.
\end{itemize}

\subsection{Visualization of Flow Prediction}\label{sec:visualization_flow_prediction}
\begin{figure}[h]
    \centering
    \includegraphics[width=\linewidth]{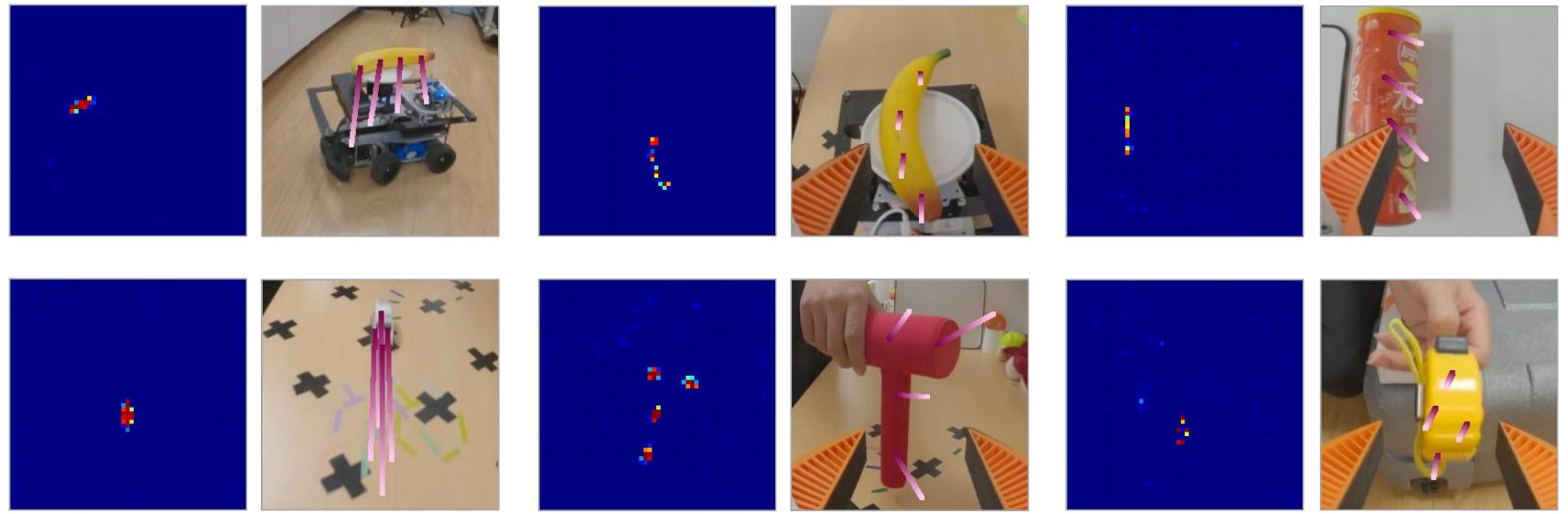}
    \caption{
    \textbf{Visualization of Flow Prediction} illustrates the pairs of heatmap and images augmented by predicted flow in different tasks. 
    }
    \label{fig:visualization_flow}
    \vspace{-5mm}
\end{figure}

A task-specific flow predictor, robust to occlusion, can be trained using under 10 minutes of video data. Figure~\ref{fig:visualization_flow} illustrates the flow generation across different tasks. In our experiments, the hyperparameters are set to four flow vectors with a 5-pixel thickness. In practice, these can be flexibly adjusted according to the dataset and the size of the manipulated objects.

\subsection{Visualization of Contrastive Learning}\label{sec:visualization_contrastive}
F2F-AP utilizes a contrastive learning objective to align the feature representation of the flow-augmented observation with that of the expected future observation.

\begin{figure}[h]
    \centering
    \includegraphics[width=0.85\linewidth]{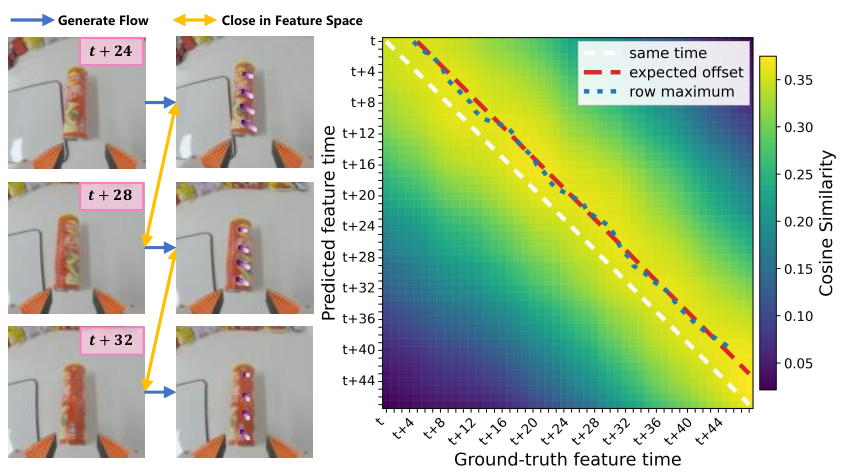}
    \caption{
    \textbf{(Right)} The heatmap indicates cosine similarity in the learned contrastive feature space, with brighter values denoting stronger alignment. The white dashed line marks same-time alignment, while the red dashed line indicates the expected temporal offset corresponding to the prediction horizon. The blue dotted line shows the empirical peak similarity for each predicted feature. \textbf{(Left)} Demonstration of how the flow-augmented observations map to the raw future observations, with the robot's egocentric observations aligned with the corresponding timestamps in the heatmap.
    }
    \label{fig:visualization_contrastive_learning}
\end{figure}

To verify whether the learned representation successfully captures the intended temporal alignment, we collect 48 consecutive samples with a time step of 0.1 seconds to compute a cosine similarity heatmap, as shown in Figure~\ref{fig:visualization_contrastive_learning}. For each sample, we compute the feature embedding of the flow-augmented image and compare it against the embeddings of future observations within the same segment. In the heatmap, each row corresponds to a flow-augmented observation feature, and each column corresponds to a ground-truth observation feature at a specific timestamp.

The resulting similarity pattern shows that the flow-augmented observation is more similar to future observations than to those at the same timestamp. This indicates that the flow-augmented images are close to future observations in the feature space, validating the design of F2F-AP by demonstrating that the policy can effectively leverage predictive feature information regarding future visual states.

\end{document}